\documentclass[10pt,journal,compsoc]{IEEEtran}
\ifCLASSOPTIONcompsoc
  \usepackage[nocompress]{cite}
\else
  \usepackage{cite}
\fi
\usepackage[utf8]{inputenc} 
\usepackage[T1]{fontenc}    
\usepackage{hyperref}       
\usepackage{url}            
\usepackage{booktabs}       
\usepackage{amsfonts}       
\usepackage{nicefrac}       
\usepackage{microtype}      
\usepackage{pifont}
\usepackage{graphicx}
\usepackage{amsmath, amsthm, amssymb, bm}
\usepackage{color, colortbl}
\definecolor{Gray}{gray}{0.97}
\usepackage{mathtools}
\usepackage{enumitem}
\usepackage{multirow}
\usepackage{bbm}
\usepackage[ruled,vlined]{algorithm2e}
\usepackage{floatrow}
\usepackage{caption}
\usepackage{booktabs}
\usepackage{rotating}
\newtheorem{prop}{Proposition}
\newtheorem{theorem}{Theorem}[section]

\newtheorem{lemma}[theorem]{Lemma}

\newtheorem{definition}{Definition}[section]

\newcommand{\base}{{\text{base}}}
\newcommand{\test}{{\text{test}}}

\newfloatcommand{capbtabbox}{table}[][\FBwidth]
\newcommand{\cmark}{\ding{51}}%
\newcommand{\xmark}{\ding{55}}%

\newcommand{\x}{\bm{x}} 
\newcommand{\w}{\bm{w}} 
\newcommand{\W}{\bm{W}} 
\newcommand{\p}{\bm{p}} 
\renewcommand{\P}{\mathbb{P}} 
\newcommand{\q}{\bm{q}} 
\newcommand{\thetab}{{\boldsymbol{\theta}}}
\newcommand{\y}{\bm{y}} 
\newcommand{\z}{\bm{z}} 
\newcommand{\abf}{\bm{a}} 
\newcommand{\bb}{\bm{b}} 
\newcommand{\E}{\mathbb{E}} 

\newcommand{\norm}[1]{\left\lVert#1\right\rVert}
\def\one{\mathbbm{1}} 
\DeclareMathOperator*{\argmax}{arg\,max}
\DeclareMathOperator*{\argmin}{arg\,min}

\begin{document}
%
\title{Mutual-Information Based \\ Few-Shot Classification}
%
%
%
%

\author{Malik~Boudiaf, Ziko~Imtiaz~Masud, Jérôme~Rony, Jose~Dolz,  Ismail~Ben~Ayed, Pablo~Piantanida
\IEEEcompsocitemizethanks{\IEEEcompsocthanksitem Malik~Boudiaf, Ziko~Imtiaz~Masud, Jérôme~Rony, José~Dolz, Ismail~Ben~Ayed are with ÉTS Montreal, Canada.\protect\\
Corresponding author: malik.boudiaf.1@etsmtl.net \protect\\

\IEEEcompsocthanksitem P. Piantanida is with Laboratoire des Signaux et Syst\`emes, CentraleSupélec CNRS Universit\'e Paris Saclay, Gif-sur-Yvette, France.
\protect\\
}

}

%
%

%

\IEEEtitleabstractindextext{
\begin{abstract}

We introduce Transductive Infomation Maximization (TIM) for few-shot learning. Our method maximizes the mutual information between the query features and their label predictions for a given few-shot task, in conjunction with a supervision loss based on the support set. We motivate our transductive loss by deriving a formal relation between the classification accuracy and mutual-information maximization. Furthermore, we propose a new alternating-direction solver, which substantially speeds up transductive inference over gradient-based optimization, while yielding competitive accuracy. We also provide a convergence analysis of our solver based on Zangwill's theory and bound-optimization arguments. 
TIM inference is modular: it can be used on top of any base-training feature extractor. Following standard transductive few-shot settings, our comprehensive experiments demonstrate that TIM outperforms state-of-the-art methods significantly across various datasets and networks, while used on top of a fixed feature extractor trained with simple cross-entropy on the base classes, without resorting to complex meta-learning schemes. It consistently brings between $2 \%$ and $5 \%$ improvement in accuracy over the best performing method, not only on all the well-established few-shot benchmarks but also on more challenging scenarios, with random tasks, domain shift and larger numbers of classes, as in the recently introduced META-DATASET. Our code is publicly available at \url{https://github.com/mboudiaf/TIM}. We also publicly release a standalone PyTorch implementation of META-DATASET, along with additional benchmarking results, at \url{https://github.com/mboudiaf/pytorch-meta-dataset}.

\end{abstract}
}

\maketitle

\IEEEdisplaynontitleabstractindextext

\IEEEpeerreviewmaketitle

\IEEEraisesectionheading{\section{Introduction}\label{sec:introduction}}

    \IEEEPARstart{D}{eep} learning models have achieved unprecedented success, approaching human-level performances when trained on large-scale labeled data. Nevertheless, the generalization of such models might be seriously challenged when dealing with new (unseen) classes, with only a few labeled instances per class. Humans, however, can learn new tasks rapidly from a handful of instances, by leveraging context and \emph{prior} knowledge. The few-shot learning (FSL) paradigm \cite{miller2000learning,fei2006one,matching_net} attempts to bridge this gap, and has recently attracted substantial research interest, with a large body of very recent works, e.g., \cite{can,dhillon2019baseline,leo,feat,liu2018learning,closer_look,team,kim2019edge,relation_net,inat_benchmark,gidaris2019boosting,prototypical_nets,maml}, among many others. In the few-shot setting, a model is first trained on labeled data with {\em base} classes. Then, model generalization is evaluated on few-shot {\em tasks}, composed of unlabeled samples from novel classes unseen during training (the {\em query} set), assuming only one or a few labeled samples (the {\em support} set) are given per novel class. 
    
    Most of the existing approaches within the FSL framework are based on the "learning to learn" paradigm or meta-learning \cite{maml,prototypical_nets,matching_net,relation_net,lee2019meta}, where the training set is viewed as a series of balanced tasks (or \textit{episodes}), to simulate test-time scenario. Popular works include prototypical networks \cite{prototypical_nets}, which describes each class with an embedding prototype and maximizes the log-probability of query samples via episodic training; matching network \cite{matching_net}, which represents query predictions as linear combinations of support labels and employs episodic training along with memory architectures; MAML \cite{maml}, a meta-learner, which trains a model to make it "easy" to fine-tune; and the LSTM meta-learner in \cite{ravi2016optimization}, which suggests optimization as a model for few-shot learning. A large body of meta-learning works followed-up lately, to only cite a few \cite{leo,tadam,Mishra18,relation_net,feat}.

    \subsection{Related work}
        \textbf{Transductive inference:} In a recent line of work, {\em transductive} inference has emerged as an appealing approach to tackle few-shot tasks \cite{dhillon2019baseline, can, kim2019edge,liu2018learning,team, nichol2018firstorder, prototype, Laplacian}, showing performance improvements over {\em inductive} inference. In the transductive setting\footnote{Transductive few-shot inference is not to be confused with semi-supervised few-shot learning \cite{tiered_imagenet, paper_to_please_R1}. The latter uses extra unlabeled data during meta-training. Transductive inference has access to exactly the same training/testing data as its inductive counterpart.}, the model classifies the unlabeled query examples of a single few-shot task at once, instead of one sample at a time as in inductive methods.
        These recent experimental observations in few-shot learning are consistent with established facts in classical transductive inference \cite{vapnik1999overview,joachim99,z2004learning}, which is well-known to outperform inductive methods on small training sets. While \cite{nichol2018firstorder} used information of unlabeled query samples via batch normalization, the authors of \cite{liu2018learning} were the first to model explicitly transductive inference in few-shot learning. Inspired by popular label-propagation concepts \cite{z2004learning}, they built a meta-learning framework that learns to propagate labels from labeled to unlabeled instances via a graph. The meta-learning transductive method in \cite{can} used attention mechanisms to propagate labels to unlabeled query samples. More closely related to our work, the recent transductive inference of Dhillion et al. \cite{dhillon2019baseline} minimizes the entropy of the network softmax predictions at unlabeled query samples, reporting competitive few-shot performances, while using standard cross-entropy training on the base classes. The competitive performance of \cite{dhillon2019baseline} is in line with several recent inductive baselines \cite{closer_look,simpleshot,tian2020rethinking}, which reported that standard cross-entropy training for the base classes matches or exceeds the performances of more sophisticated meta-learning procedures. Also, the performance of \cite{dhillon2019baseline} is in line with established results in the context of semi-supervised learning, where entropy minimization is widely used \cite{grandvalet2005semi,miyato2018virtual,berthelot2019mixmatch}. It is worth noting that the inference runtimes of transductive methods are, typically, much higher than their inductive counterparts. For, instance, the authors of \cite{dhillon2019baseline} fine-tune all the parameters of a deep network during inference, which is several orders of magnitude slower than inductive methods such as ProtoNet \cite{prototypical_nets}. Also, based on matrix inversion, the transductive inference in \cite{liu2018learning} has a complexity that is cubic in the number of query samples.\\

        \noindent\textbf{Info-max principle:} While the semi-supervised and few-shot learning works in \cite{grandvalet2005semi,dhillon2019baseline} build upon Barlow's principle of entropy minimization \cite{Barlow1989Unsupervised}, our few-shot formulation is inspired by the general info-max principle enunciated by Linsker \cite{Linsker1988Self}, which formally consists in maximizing the Mutual Information (MI) between the inputs and outputs of a system. In our case, the inputs are the query features and the outputs are their label predictions. The idea is also related to info-max in the context of clustering \cite{clustering_infomax,HuICML17,jabi}. More generally, info-max principles, well-established in the field of communications, were recently used in several deep-learning problems, e.g., representation learning \cite{deep_infomax,cpc}, metric learning \cite{boudiaf2020metric} or domain adaptation \cite{liang2020we}, among other works.

        
    
    
    \subsection{Contributions}
        \begin{itemize}
            \item We propose Transductive Information Maximization (TIM) for few-shot learning.  Our method maximizes the MI between the query features and their label predictions for a few-shot task at inference, while minimizing the cross-entropy loss on the support set. We formally motivate the mutual information loss as a surrogate of the classification error.
            \item We derive an alternating-direction solver for our loss, which substantially speeds up transductive inference over gradient-based optimization, while yielding competitive accuracy. Furthermore, we provide a convergence analysis based on Zangwill's theory and bound-optimization arguments. 
            \item Following standard transductive few-shot settings, our comprehensive evaluations show that TIM outperforms state-of-the-art methods substantially across various datasets and networks,  while using a simple cross-entropy training on the base classes, without complex meta-learning schemes. It consistently brings between $2 \%$ and $5 \%$ of improvement in accuracy over the best performing method, not only on all the well-established few-shot benchmarks but also on more challenging, recently introduced scenarios, with domain shifts and larger numbers of ways.
        \end{itemize}

This work extends and generalizes in many different ways our preliminary results in \cite{boudiaf2020transductive}, published at the NeurIPS 2020 conference. More specifically, it introduces an information-theoretic justification for the previous formulation in \autoref{sec:mi_justif}, it provides new results on the convergence of our TIM-ADM algorithm in \autoref{sec:adm} and \autoref{sec:adm_convergence}, and reports several new experiments and benchmarking results on META-DATASET, a recently introduced, challenging few-shot dataset, in \autoref{sec:beyond_standard_benchmarks}.

\section{Transductive Information Maximization}

    \subsection{Few-shot setting}
    	Assume we are given a labeled training set, $\mathcal{X}_\base\coloneqq \{\x_i, \y_i\}_{i=1}^{N_\base}$, where 
    	$\x_i$ denotes raw features of sample $i$ and $\y_i$ its associated one-hot encoded label. Such labeled set is often referred to as the {\em meta-training} or {\em base} dataset in the few-shot literature. Let  
    	$\mathcal{Y}_\base$ denote the set of classes for this base dataset. The few-shot scenario assumes that we are given a {\em test} dataset: $\mathcal{X}_\test\coloneqq\{\x_i, \y_i\}_{i=1}^{N_\test}$, with a completely new set of classes $\mathcal{Y}_{\test}$ such that $\mathcal{Y}_\base \cap \mathcal{Y}_\test = \emptyset$, from which we create randomly sampled few-shot {\em tasks}, each with a few labeled examples. \\
    	
        \noindent
        \textbf{Standard tasks:} Traditionally, models are (trained and) evaluated on $K$-ways $N_{S}$-shot task, which involve randomly sampling $N_S$ labeled examples from each of $K$ different classes, also chosen at random. Let $\mathcal{S}$ denote the set of these labeled examples with size $|\mathcal{S}|=N_{{S}}.K$, referred to as the \textit{support} set. . Furthermore, each task has a {\em query} set denoted by $\mathcal{Q}$ composed of $ | \mathcal{Q}| =N_{{Q}}.K$ unlabeled (unseen) examples from each of the $K$ classes. With models trained on the base set, few-shot techniques use the labeled support sets to adapt to the tasks at hand, and are evaluated based on their performances on the unlabeled query sets. \\
        
        \noindent
        \textbf{Random tasks:} Recently, there has been an increasing interest to move towards random tasks, which arguably provide a more challenging but more realistic scenario. In particular, \textsc{Meta-dataset} \cite{triantafillou2019meta} proposes several improvements over the standard setting: break the symmetry in the support set by having each class contain a different random number of labelled samples, randomly sample the total number of support samples for a task and randomly samples the total number of ways. Both standard and random task setting will be evaluated in \autoref{sec:experiments}.

	\subsection{Proposed formulation}\label{sec:formulation}
	
    	We begin by introducing some basic notations and definitions before presenting our overall Transductive Information Maximization (TIM) loss and the different optimization strategies for tackling it. For a given $K$-way few-shot task, with a support set $\mathcal{S}$ and a query set $\mathcal{Q}$, let $X$ denote the random variable associated with the raw features within ${\mathcal{S} \cup \mathcal{Q}}$, and let $Y \in  \mathcal{Y}=\{1, \dots, K \}$ be the random variable associated with the data labels. Let $f_{\boldsymbol{\phi}}: \mathcal{X} \longrightarrow \mathcal{Z} \subset \mathbb{R}^d$ denote the encoder (\emph{i.e.}, feature-extractor) function of a deep neural network, where $\boldsymbol{\phi}$ denotes the trainable parameters, and $\mathcal{Z}$ stands for the set of embedded features. The encoder is first trained from the base training set $\mathcal{X}_\base$ using the standard cross-entropy loss, without any meta training or specific sampling schemes. Then, for each specific few-shot task, we propose to minimize a mutual-information loss defined over the query samples.
    	
    	Formally, we define a soft-classifier associated to the random variable $\widehat{Y} \in \mathcal{Y}$ and parametrized by weight matrix $\W = [\w_1, \dots, \w_K] \in \mathbb{R}^{K \times d}$, whose posterior distribution over labels given features\footnote{In order to simplify our  notations, we deliberately omit the dependence of posteriors $p_{ik}$ on the network parameters $(\boldsymbol{\phi},\W)$. Also, $p_{ik}$ takes the form of \emph{softmax} predictions, but we omit the normalization constants.}, ${p_{ik} \coloneqq \mathbb{P}(\widehat{Y}=k|{ X=\x_i}; \W, \boldsymbol{\phi})}$, and marginal distribution over query labels, ${\widehat{p}_k \coloneqq \mathbb{P}(\widehat{Y}_\mathcal{Q}=k; \W, \bm{\phi})}$, are given by:
    	\begin{equation}\label{eq:distance_based_classifier}
    		p_{ik} \propto \exp\left(-\frac{\tau}{2}\norm{\w_k - \z_i}^2\right),~ \textrm{ and } ~ \widehat{p}_k = \frac{1}{|\mathcal{Q}|} \sum_{i \in \mathcal{Q}} p_{ik}, \,  
    	\end{equation}
    	where ${z_i=f_{\boldsymbol{\phi}}(\x_i) / \norm{f_{\boldsymbol{\phi}}(\x_i)}_2}$ the L2-normalized embedded features, and $\tau$ is a temperature parameter. 
    
         Now, for each single few-shot task, we introduce our empirical weighted mutual information between the query samples and their latent labels, which integrates two terms: The first is an empirical (Monte-Carlo) estimate of the conditional entropy of labels given the query raw features, denoted $\mathcal{H}(\widehat{Y}_\mathcal{Q}|X_\mathcal{Q})$, while the second is the empirical label-marginal entropy, $\mathcal{H}(\widehat{Y}_\mathcal{Q})$:
        \begin{align}
        \label{prior-aware-MI}
            \begin{split}
                &\mathcal{I}_{\alpha}(X_\mathcal{Q}; \widehat{Y}_\mathcal{Q}) \coloneqq  \\
               & \underbrace{-\sum_{k=1}^K \widehat{p}_{k} \log \widehat{p}_{k}}_{\mathcal{H}(\widehat{Y}_\mathcal{Q}):~\text{marginal entropy}} + \alpha  \underbrace{\frac{1}{|\mathcal{Q}|} \sum_{i \in \mathcal{Q}}\sum_{k=1}^K p_{ik} \log (p_{ik})}_{-\mathcal{H}(\widehat{Y}_\mathcal{Q}| X_\mathcal{Q}):~\text{conditional entropy}}
            \end{split}
        \end{align}
        with $\alpha$ a non-negative hyper-parameter. Notice that setting $\alpha=1$ recovers the standard mutual information. Setting $\alpha<1$ allows us to down-weight the conditional entropy term, whose gradients may dominate the marginal entropy gradients as the predictions move towards the vertices of the simplex. The role of both terms in Eq. \eqref{prior-aware-MI} will be discussed after introducing our overall transductive inference loss in the following, by embedding supervision from the task's support set. 

        We embed supervision information from support set $\mathcal{S}$ by integrating a standard cross-entropy loss $\textrm{CE}$ with the information measure in Eq. \eqref{prior-aware-MI}, which enables us to formulate our Transductive Information Maximization (\textbf{TIM}) loss as follows:
        	\begin{align}\label{eq:tim_objective}
        	    \begin{split}
                    \min_{\W} ~ \mathcal{L} &=~ \lambda \cdot \textrm{CE} -   \mathcal{I}_{\alpha}(X_\mathcal{Q}; \widehat{Y}_\mathcal{Q}) \\
        	        \textrm{ with } \quad \textrm{CE} &\coloneqq -\frac{1}{|\mathcal{S}|} \sum_{i \in \mathcal{S}}\sum_{k=1}^K y_{ik} \log (p_{ik}),
        	    \end{split}
        	\end{align}
        where $y_{ik}$ denotes the $k^{th}$ component of the one-hot encoded label $\y_i$ associated to the $i$-th support sample. Non-negative hyper-parameters $\alpha$ and $\lambda$ will be fixed to $\alpha=\lambda=0.1$ in all our experiments. It is worth to discuss in more details the role (importance) of the mutual information terms in (\ref{eq:tim_objective}): 
        
    	\begin{itemize}[leftmargin=*]
    	    \item Conditional entropy $\mathcal{H}(\widehat{Y}_\mathcal{Q}|X_\mathcal{Q})$ aims at minimizing the uncertainty of the posteriors at unlabeled query samples, thereby encouraging the model to output {\em confident} predictions\footnote{The global minima of each pointwise entropy in the sum of $\mathcal{H}(\widehat{Y}_\mathcal{Q}|X_\mathcal{Q})$ are one-hot vectors at the vertices of the simplex.}. This entropy loss is widely used in the context of semi-supervised learning (SSL) \cite{grandvalet2005semi,miyato2018virtual,berthelot2019mixmatch}, as it models effectively the {\em cluster} assumption: The classifier's boundaries should not occur at dense regions of the unlabeled features \cite{grandvalet2005semi}. Recently, \cite{dhillon2019baseline} introduced this term for few-shot learning, showing that entropy fine-tuning on query samples achieves competitive performances. In fact, if we remove the marginal entropy $\mathcal{H}(\widehat{Y}_\mathcal{Q})$ in objective \eqref{eq:tim_objective}, our TIM objective reduces to the loss in \cite{dhillon2019baseline}. The conditional entropy $\mathcal{H}(\widehat{Y}_\mathcal{Q}|X_\mathcal{Q})$ is of paramount importance but its optimization  requires special care, as its optima may easily lead to degenerate (non-suitable) solutions on the simplex vertices, mapping all samples to a single class. Such care may consist in using small learning rates and fine-tuning the whole network (which itself often contains several layers of regularization) as done in \cite{dhillon2019baseline}, both of which significantly slow down transductive inference.
    	    
    	    \item The label-marginal entropy regularizer $\mathcal{H}(\widehat{Y}_\mathcal{Q})$ encourages the marginal distribution of labels to be uniform, thereby avoiding degenerate solutions obtained when solely minimizing conditional entropy. Hence, it is highly important as it removes the need for implicit regularization, as mentioned in the previous paragraph. In particular, high-accuracy results can be obtained even using higher learning rates and fine-tuning only a fraction of the network parameters (classifier weights $\W$ instead of the whole network), speeding up substantially transductive runtimes. As it will be observed from our experiments, this term brings substantial improvements in performances (e.g., up to $10\%$ increase in accuracy over entropy fine-tuning on the standard few-shot benchmarks), while facilitating optimization, thereby reducing transductive runtimes by orders of magnitude.

    	\end{itemize}

    	\subsection{Mutual information and risk}\label{sec:mi_justif}

            We now give some theoretical justification on the proposed formulation, especially on the mutual information $\mathcal{I}_{\alpha}$ used in Eq. \eqref{eq:tim_objective}. First, let us make a subtle difference between the soft decision $\widehat{Y}|X$ that would correspond to sampling a decision from the softmax distribution output by the network, and the hard-decision:  
            \begin{align}
                \widehat{y}^\star(X) \coloneqq \argmax_{y \in \mathcal{Y}} ~ \P(\widehat{Y}=y|X)
            \end{align}
            that simply picks the class with the highest softmax score. Let us now introduce define the probability of classification error (or risk) as:
            \begin{align}
                \P_e \coloneqq \P\left(Y_\mathcal{Q} \ne \widehat{y}^\star(X_\mathcal{Q})\right),    
            \end{align}
            where recall $X_\mathcal{Q},Y_\mathcal{Q}$ model the data distribution on the query set. \\
            
            We argue that without any assumption, mutual information needs not be a well-suited criterion for classification purposes. To illustrate this point, consider any permutation (except the identity) of the labels $\pi: \mathcal{Y} \rightarrow \mathcal{Y}$, and a classifier $\widehat{Y}|X$ such that $\P(\widehat{Y}=\pi(y)|X=\mathbf{x}, Y=y)=1$. Then, one can verify that the classifier $\widehat{Y}|X$ satisfies both 100\% classification error and maximum mutual information $\mathcal{I}(X; \widehat{Y}) = \log(|\mathcal{Y}|)$. Therefore, restricting assumptions on the classifier $\widehat{Y}|X$ must apply in order to relate the mutual information objective to the probability of classification error. In this section, we address the following question: \textit{Can we find sufficient conditions on the classifier $\widehat{Y}_\mathcal{Q}|X_\mathcal{Q}$ such that the mutual information $\mathcal{I}(X_\mathcal{Q};\widehat{Y}_\mathcal{Q})$ and the classification error $\P_e$ can be explicitly related ?}

            In our following result, we draw a theoretical link between mutual information maximization and classification error. Specifically, under the assumption that a classifier's confusion matrix is diagonal dominant, we show its risk can be upper bounded by a non-decreasing function of mutual information.
            
            \begin{prop} \label{prop:mi_theory}
                Consider the classifier $\widehat{Y}_\mathcal{Q}|X_\mathcal{Q}$ defined on the query set $\mathcal{Q}$. Assume the confusion matrix of $\widehat{Y}_\mathcal{Q}|X_\mathcal{Q}$ is diagonal dominant that is:
                \begin{align*}
                    \P(\widehat{Y}_\mathcal{Q}=y,Y_\mathcal{Q}=y) > \P(\widehat{Y}_\mathcal{Q}=y^\prime,Y_\mathcal{Q}=y), \, \,  \forall \, \, ~ y \neq y^\prime.
                \end{align*}
                Without loss of generality, we assume there exists $\epsilon > 0$ such that:
                \begin{align*}
                    \P(\widehat{Y}_\mathcal{Q}=y|Y_\mathcal{Q}=y) > 1 - \epsilon, \quad \forall ~ y \in \mathcal{Y}.
                \end{align*}
                Then the following relation holds:
                \begin{align} \label{eq:general_pe_upper_bound}
                    \P_e \leq & ~
                        \delta \left(\sqrt{\frac{\ln 2}{2} \mathcal{D}_{\text{KL}} \big (\widehat{Y}_\mathcal{Q} \| Y_\mathcal{Q}\big)}\right) + \delta\left(\mathcal{H}(\widehat{Y}_\mathcal{Q}|X_\mathcal{Q})\right) + g(\epsilon),
                \end{align}
                where $\delta(\cdot)$ is a strictly increasing function on the restricted domain $\left[0, \frac{|\mathcal{Y}| - 1}{|\mathcal{Y}} \right]$, with $\delta(0)=0$ and $f(0)=0$.
                
                In the case of a uniform prior distribution over classes $Y \sim  \mathcal{U} [1:K]$, expression \eqref{eq:general_pe_upper_bound} becomes:
                \begin{align}\label{eq:uniform_pe_upper_bound}
                    \P_e \leq & ~
                        \delta \left(\sqrt{\frac{\ln 2}{2} \big [ \log(K) - \mathcal{H}(\widehat{Y}_\mathcal{Q}) \big] }\right)\nonumber \\& + \delta\left(\mathcal{H}(\widehat{Y}_\mathcal{Q}|X_\mathcal{Q})\right) + g(\epsilon).
                \end{align}
            \end{prop}
            The full proof of Proposition \ref{prop:mi_theory} is provided in the Supplemental. In all few-shot benchmarks, we consider a uniform distribution on the query set. Therefore, Eq. \eqref{eq:uniform_pe_upper_bound} holds on the query set, which clearly motivates the transductive mutual information. 
            
\begin{figure}
                \centering
                \includegraphics[width=0.9\textwidth]{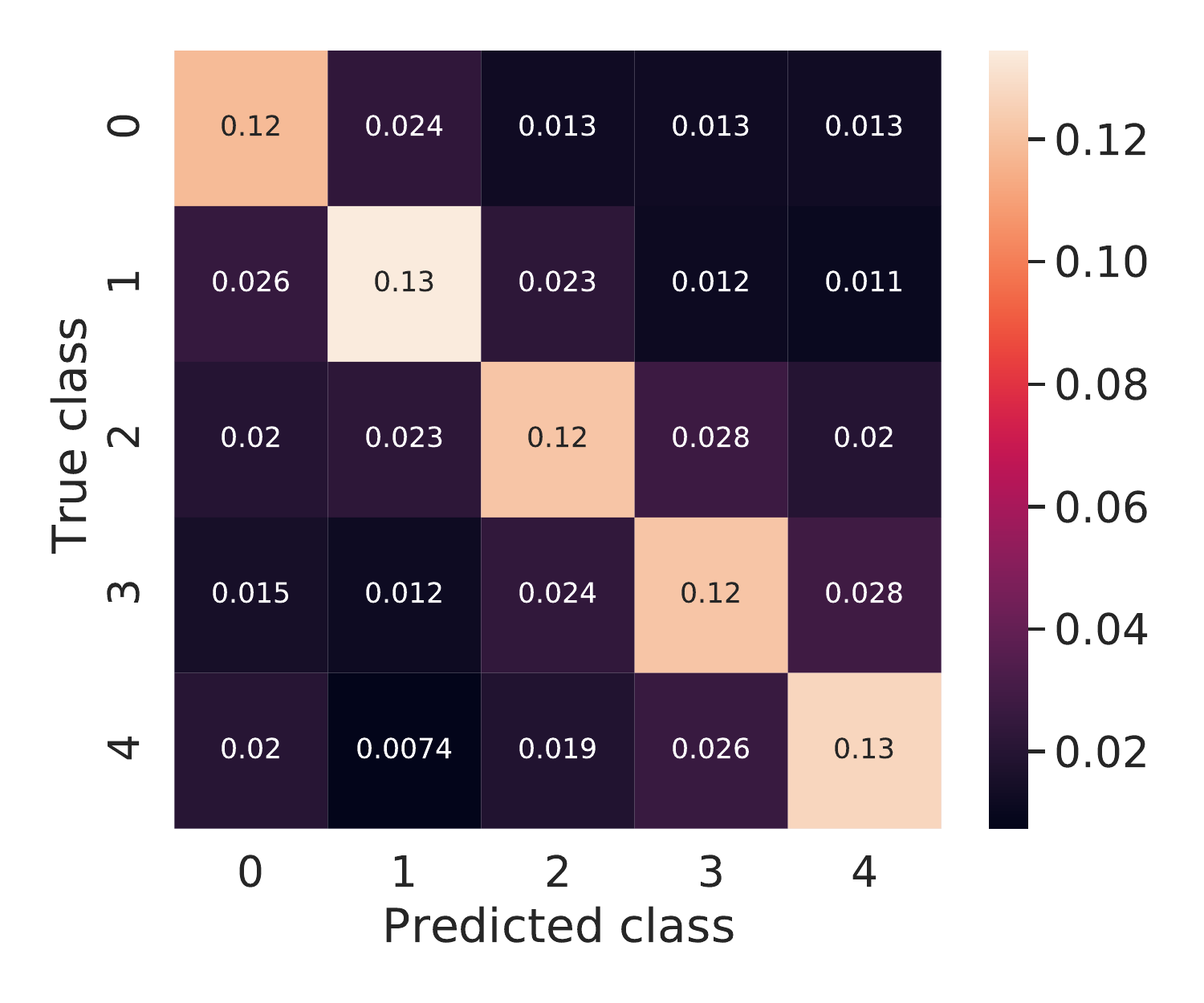}
                \caption{Normalized confusion matrix of the classifier at initialization (t=0), averaged over 1000 5-ways 1-shot tasks randomly sampled from \textit{mini}-ImageNet. The assumption of diagonal dominance in Proposition \ref{prop:mi_theory} is verified on average.}
                \label{fig:confusion_matrix}
            \end{figure}

            In the case of a perfectly diagonal confusion matrix, i.e., $\epsilon=0$, one can verify that a maximum mutual information, i.e. $\mathcal{H}(\widehat{Y}_\mathcal{Q})=\log(K)$ and $\mathcal{H}(\widehat{Y}_\mathcal{Q}|X_\mathcal{Q})=0$ leads to a perfect classification $\P_e=0$. In practice, such assumption is surely unrealistic, but we show on \autoref{fig:confusion_matrix} that the assumption of a diagonal dominant confusion matrix is verified on average even at initialization.
            
\section{Optimization}\label{sec:optim}
    At this stage, we consider that the feature extractor has already been trained on base classes (using standard cross-entropy). We now propose two methods for minimizing our objective (\ref{eq:tim_objective}) for each test task. The first one is based on standard Gradient Descent (GD). The second is a novel way of optimizing mutual information, and is inspired by the Alternating Direction Method of Multipliers (ADMM). For both methods:
    \begin{itemize}
        \item The pre-trained feature extractor $f_\phi$ is kept fixed. Only the weights $\W$ are optimized for each task. Such a choice is discussed in details in \autoref{sec:ablation_terms}. Overall, and interestingly, we found that fine-tuning only classifier weights $\W$, while fixing feature-extractor parameters $\phi$, yielded the best performances for our mutual-information loss.
        \item For each task, weights $\W$ are initialized as the class prototypes of the support set: 
        \begin{align} \label{eq:weight_init}
            \w_k^{(0)}= \frac{\sum_{i \in \mathcal{S}}y_{ik} \z_i}{\sum_{i \in \mathcal{S}}y_{ik}}.
        \end{align}
    \end{itemize}

    \subsection{Gradient descent (TIM-GD)} 
        A straightforward way to minimize our loss in Eq. (\ref{eq:tim_objective}) is to perform gradient descent over $\W$, which we update using all the samples from the few-shot task (both support and query) at once (i.e., no mini-batch sampling). This gradient approach yields our overall best results, while being one order of magnitude faster than the transductive entropy-based fine-tuning in \cite{dhillon2019baseline}. As will be shown later in our experiments, the method in \cite{dhillon2019baseline} needs to fine-tune the whole network (i.e., to update both $\phi$ and $\W$), which provides implicit regularization, avoiding the degenerate solutions of entropy minimization. However, TIM-GD (with $\W$-updates only) still remains two orders of magnitude slower than inductive closed-form solutions \cite{prototypical_nets}. In the following, we present a more efficient solver for our problem. The algorithm associated to TIM-GD is presented in Algorithm \ref{algo:tim_gd}.

    \subsection{Alternating direction method (TIM-ADM)} \label{sec:adm}
        We derive an Alternating Direction Method (ADM) for minimizing our objective in \eqref{eq:tim_objective}. Such scheme yields substantial speedups in transductive learning (one order of magnitude), while maintaining excellent accuracy performances. To do so, we introduce auxiliary variables representing latent assignments of query samples, and minimize a mixed-variable objective by alternating two sub-steps, one optimizing w.r.t classifier's weights $\W$, and the other w.r.t the auxiliary variables $\q$. 

	    \begin{prop}\label{prop:mi_adm}
	        The objective $\mathcal{L}$ in Eq. \eqref{eq:tim_objective} can be minimized via the following constrained formulation of the problem:
	        \begin{align}\label{eq:mi_adm}
                \min_{\W, \q} \,\,  & ~\mathcal{L}_{ADM} ~=~\underbrace{-\frac{\lambda}{|\mathcal{S}|} \sum_{i \in \mathcal{S}} \y_{i}^T \log (\p_{ik})}_{\mathrm{CE}} 
                + \underbrace{\sum_{k=1}^K \widehat{q}_{k}\log {\widehat{q}_{k}}}_{\sim \mathcal{H}(\widehat{Y}_\mathcal{Q}) } \\ 
                & \qquad \qquad \quad \underbrace{-\frac{\alpha}{|\mathcal{Q}|} \sum_{i \in \mathcal{Q}}\q_{i}^T \log (\p_{i})}_{\sim \, \mathcal{H}(\widehat{Y}_\mathcal{Q}|X_\mathcal{Q})} 
                + \underbrace{\frac{\beta}{|\mathcal{Q}|} \sum_{i \in \mathcal{Q}} \q_{i}^T\log \frac{\q_{i}}{\p_{i}}}_{\text{Penalty} \, \equiv\, \mathcal{D}_{\mathrm{KL}}(\q \| \p )}
                \nonumber \\
                \text{s.t}  \quad & \sum_{k=1}^K q_{ik}=1, \quad q_{ik} \geq 0, \quad i \in \mathcal{Q}, \quad k \in \{1, \dots, K\}, \nonumber
	        \end{align}
        where ${\q=[q_{ik}] \in \mathbb R^{|\mathcal{Q}| \times K}}$ are auxiliary variables,  ${\textbf{p}=[p_{ik}] \in \mathbb R^{|\mathcal{Q}| \times K}}$, $\widehat{q}_k=\frac{1}{|\mathcal{Q}|}\sum\limits_{i \in \mathcal{Q}}q_{ik}$, and $\beta > 0$ a Lagrangian multiplier.
        \end{prop}
	    
	    \begin{IEEEproof}
	    It is straightforward to notice that, when equality constraints $q_{ik} = p_{ik}$ are satisfied, the last term in objective (\ref{eq:mi_adm}), which can be viewed as a soft penalty for enforcing those equality constraints,  vanishes. Objectives (\ref{eq:tim_objective}) and (\ref{eq:mi_adm}) then become equivalent.
	    \end{IEEEproof}
	    
        Splitting the problem into sub-problems on $\W$ and $\q$ as in Eq. \eqref{eq:mi_adm} is closely related to the general principle of ADMM (Alternating Direction Method of Multipliers) \cite{boyd}, except that the KL divergence is not a typical penalty for imposing the equality constraints\footnote{Typically, ADMM methods use multiplier-based quadratic penalties for enforcing the equality constraint.}. Note that the multiplier $\beta$ is kept fixed in practice, and treated as an hyperparameter. The main idea is to \textbf{decompose the original problem into two easier sub-problems}, one over $\W$ and the other over $\q$, which can be alternately solved, each in closed-form. Interestingly, this KL penalty is important as it completely removes the need for dual iterations for the simplex constraints in Eq. \eqref{eq:mi_adm}, yielding closed-form solutions. 
		    
	    We now describe the TIM-ADM algorithm. Consider the following closed-form updates for $t>0$:
        \begin{align}
            q_{ik}^{(t+1)} &\propto \quad \displaystyle \frac{ \left ( p_{ik}^{(t)} \right )^{1+\frac{\alpha}{\beta}}}{\displaystyle \left ( \sum_{i \in \mathcal{Q}} \left ( p_{ik}^{(t)} \right )^{1+\frac{\alpha}{\beta}} \right )^{\frac{1}{1+\beta}}},  \label{eq:q-update}\\
            \w_k^{(t+1)} &\leftarrow \frac{\displaystyle\frac{\lambda}{\beta+\alpha} \displaystyle\sum\limits_{i \in \mathcal{S}} f(y_{ik})  +  \frac{|\mathcal{S}|}{|\mathcal{Q}|} \displaystyle\sum\limits_{i \in \mathcal{Q}} f(q_{ik}^{(t+1)})}{\displaystyle\frac{\lambda}{\beta+\alpha} \displaystyle\sum\limits_{i \in \mathcal{S}} y_{ik} +  \frac{|\mathcal{S}|}{|\mathcal{Q}|} \sum\limits_{i \in \mathcal{Q}} q_{ik}^{(t+1)}}, \label{eq:w-update}
	    \end{align}
        where $f(y) =  y ~\z_i + p_{ik}^{(t)} (\w_k^{(t)} - \z_i)$, and where we recall $\propto$ means "proportional to" (with the correct constant s.t.  $\sum_k q_{ik}=1$).

\begin{figure}
            \centering
            \caption{Maximum eigenvalues of hessian matrices $H_k^{\mathcal{S}}$ and $H_k^{\mathcal{Q}}$ introduced in Proposition \ref{prop:adm_decreases_loss} averaged over 1000 5-ways 1-shot \textit{mini}-ImageNet tasks. (Left) $\gamma_{\mathcal{S}}^{max}$ represents the maximum eigenvalue of $H_k^{\mathcal{S}}$, over all classes $k$. (Right) $\gamma_{\mathcal{Q}}^{max}$ represents the maximum eigenvalue of $H_k^{\mathcal{Q}}$, over all classes $k$.}
            \label{fig:eigenvalues}.
            \includegraphics[width=\textwidth]{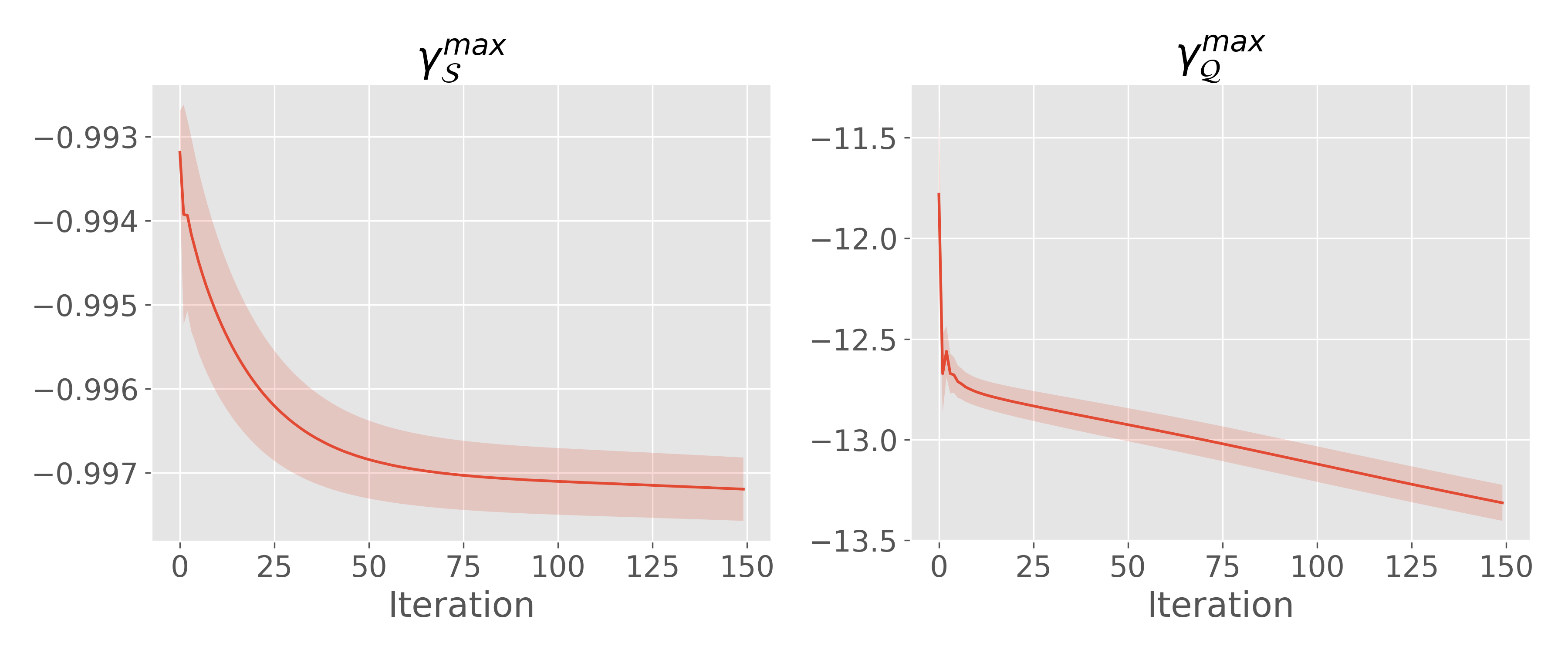}
        \end{figure}

	    \begin{prop}\label{prop:adm_decreases_loss}
            Assume that at each iteration $t > 0$, and for each class $k \in \{1, \dots, K\}$, the matrices:
            \begin{align*}
                H_k^{\mathcal{S}} = \sum_{i \in \mathcal{S}} p_{ik}^{(t)}(p_{ik}^{(t)} - 1) (\z_i - \w_k^{(t)})(\z_i - \w_k^{(t)})^T - p_{ik}^{(t)}~\textrm{Id}, \\
                H_k^{\mathcal{Q}} = \sum_{i \in \mathcal{Q}} p_{ik}^{(t)}(p_{ik}^{(t)} - 1) (\z_i - \w_k^{(t)})(\z_i - \w_k^{(t)})^T - p_{ik}^{(t)}~\textrm{Id}
            \end{align*}
            are both semi-definite-negative, where $\textrm{Id} \in \mathbb R^{d \times d}$ is the identity matrix. Then ADM formulation in Proposition \ref{prop:mi_adm} can be minimized w.r.t auxiliary assignment variables $\q$ and classifier weights $\W$ by alternating the closed-form updates \eqref{eq:q-update} and \eqref{eq:w-update}. Specifically, updates \eqref{eq:q-update} and \eqref{eq:w-update} for some $t>0$ are guaranteed to fulfill:
	        \begin{align}
	            \mathcal{L}_{\textrm{ADM}}(\q^{(t+1)}, \W^{(t+1)}) \leq \mathcal{L}_{\textrm{ADM}}(\q^{(t)}, \W^{(t)}).
	        \end{align}
	    \end{prop}
        \begin{IEEEproof}
		    A detailed proof is deferred to the supplementary material. Here, we summarize the main technical ingredients. Keeping the auxiliary variables $\q$ fixed, we optimize
		    an auxiliary bound on Eq. \eqref{eq:mi_adm}, that is convex w.r.t $\W$. With $\W$ fixed, the objective \eqref{eq:mi_adm} is strictly convex w.r.t the auxiliary variables $\q$ whose updates come from a closed-form solution of the KKT (Karush–Kuhn–Tucker) conditions. Interestingly, the negative entropy of auxiliary variables, which appears in the penalty term, handles implicitly the simplex constraints, which removes the need for dual iterations to solve the KKT conditions.
	    \end{IEEEproof}
        In Proposition \ref{prop:adm_decreases_loss}, the symmetric matrices $H_k^{\mathcal{S}}$ and $H_k^{\mathcal{Q}}$ introduced corresponds to hessian matrices w.r.t parameters $\w_k$, and their semi-definite negativeness allow to interpret $\W$-update \eqref{eq:w-update} as a bound optimization step. This assumption is empirically verified, as shown in \autoref{fig:eigenvalues}. The algorithm associated to TIM-ADM is presented in Algorithm \ref{algo:tim_adm}. Note the loss $\mathcal{L}_{ADM}$ is bounded from below, as the cross-entropy $\textrm{CE}$ is positive, the two entropy terms are bounded between 0 and $\log(K)$, and the $\mathcal{D}_{KL}$ is positive. Therefore, Proposition \ref{prop:adm_decreases_loss} allows us to affirm that, provided the assumptions are respected, the sequence of loss values $\{\mathcal{L}_{ADM}(\W^{(t)}, \q^{(t)})\}_{t \in \mathbb N}$ is both non-increasing and bounded from below, hence converges. However, this does not inform us on the behavior of the parameter sequence $\{\W^{(t)}, \q^{(t)}\}_{t \in \mathbb N}$. The latter is examined in the next \autoref{sec:adm_convergence}.
                
        \begin{algorithm}[t]
            \SetKwInOut{Input}{Input}\SetKwInOut{Output}{output}
            \SetAlgoLined
            \Input{ Pre-trained encoder $f_{\bm{\phi}}$, Task $\{\mathcal{S}, \mathcal{Q}\}$,\\ \# iterations $iter$, Temperature $\tau$, \\Weights $\{\lambda, \alpha\}$} \vspace{0.5em}
            $\z_i \leftarrow f_{\bm{\phi}}(\x_i) /\norm{f_{\bm{\phi}}(\x_i)}_2$ , $i \in \mathcal{S} \cup \mathcal{Q}$ \\
            Initialize $\w_k^{(0)},~ k \in \{1, \dots, K\}$ with Eq. \eqref{eq:weight_init}  \\
            \For{$t\leftarrow 0$ \KwTo $iter$}{
                Compute $\p^{(t)}$ with Eq. \eqref{eq:distance_based_classifier} \\
                Update $\q^{(t+1)}$ with Eq. \eqref{eq:q-update} \\
                  Update $\w_k^{(t+1)}$ with Eq. \eqref{eq:w-update}
                
            }
            \KwResult{Query predictions $\widehat{Y}_{i} = \argmax_{k} p_{ik}$ , $i \in \mathcal{Q}$}
            \caption{TIM-ADM inference}
            \label{algo:tim_adm}
        \end{algorithm}

        \begin{algorithm}[t]
            \SetKwInOut{Input}{Input}\SetKwInOut{Output}{output}
            \SetAlgoLined
            \Input{ Pre-trained encoder $f_{\bm{\phi}}$, Task $\{\mathcal{S}, \mathcal{Q}\}$, \\ 
            \# iterations $iter$, Temperature $\tau$, \\ Weights $\{\lambda, \alpha\}$, Learning rate $\mu$} \vspace{0.5em}
            $\z_i \leftarrow f_{\bm{\phi}}(\x_i) /\norm{f_{\bm{\phi}}(\x_i)}_2$ , $i \in \mathcal{S} \cup \mathcal{Q}$ \\
            Initialize $\w_k^{(0)},~ k \in \{1, \dots, K\}$ with Eq. \eqref{eq:weight_init}  \\
            \For{$t\leftarrow 0$ \KwTo $iter$}{
                Compute $\p^{(t)}$ with Eq. \eqref{eq:distance_based_classifier}
      	        $\w_k^{(t+1)} \leftarrow \w_k^{(t)} - \mu \nabla_{\w_k} \mathcal{L}$
                
            }
            \KwResult{Query predictions $\widehat{Y}_{i} = \argmax_{k} p_{ik}$ , $i \in \mathcal{Q}$}
            \caption{TIM-GD inference}
            \label{algo:tim_gd}
        \end{algorithm}
        
    \subsection{Convergence of TIM-ADM} \label{sec:adm_convergence}
    
        In this section, we study the convergence of the sequence $\{\W^{(t)}, \q^{(t)}\}_{t \in \mathbb N}$ in the TIM-ADM method. The idea is to show that each $\q$-update and each $\W$-udpate each strictly decrease the objective function unless the method has reached a stationary point. To formalize this idea, we analyze our proposed TIM-ADM algorithm through the lens of Zangwill’s global convergence theory \cite{zangwill1969nonlinear}, which provides a simple but general framework to study the convergence of iterative algorithms. Note that this theory was already used to prove the convergence of the concave-convex \cite{sriperumbudur2009convergence} and the EM/GEM iterative procedures \cite{wu1983convergence}. In particular, we show that all limits points of any sequence $\{\W^{(t)}, \q^{(t)}\}_{t \in \mathbb N}$ produced by our algorithm are stationary points. To avoid interrupting the flow of the main paper, we hereby only provide our convergence result. We defer the technical background on convergence of iterative algorithms that leads to our main result to the supplementary material. 
    
        
        \begin{prop} \label{prop:tim_adm_convergence}
            Starting from any initial point $\{\W^{(0)}, \q^{(0)}\}$ such that $\mathcal{L}_{ADM}(\W^{(0)}, \q^{(0)}) < \infty$, and assuming that at every iteration $t>0$, the assumptions of Proposition \ref{prop:adm_decreases_loss} hold, then every limit point $\{\W^{*}, \q^{*}\}$ of the sequence $\{\W^{(t)}, \q^{(t)}\}_{t \in \mathbb N}$ built using TIM-ADM (Algorithm \ref{algo:tim_adm}) is a fixed point (running more iterations leaves $\{\W^{*}, \q^{*}\}$ unchanged).
        \end{prop}
        
        The full proof of Proposition \ref{prop:tim_adm_convergence} is provided in the Supplemental material. The proof is a direct application of Zangwill's convergence theorem.

\section{Experiments} \label{sec:experiments}

    \begin{table*}
    \centering
    \small
    \caption{Comparison to the state-of-the-art methods on \textit{mini}-ImageNet, \textit{tiered}-Imagenet and CUB. The methods are sub-grouped into transductive and inductive methods, as well as by backbone architecture. Our results (gray-shaded) are averaged over 10,000 episodes. "-" signifies the result is unavailable.}
    \begin{tabular}{lccccccccc}
        \toprule
        & & &\multicolumn{2}{c}{\textbf{\textit{mini}-ImageNet}} & \multicolumn{2}{c}{\textbf{\textit{tiered}-ImageNet}} & \multicolumn{2}{c}{\textbf{CUB}} \\
         Method & Transd. & Backbone & 1-shot & 5-shot & 1-shot & 5-shot & 1-shot & 5-shot\\
         \midrule
         MAML \cite{maml}& \multirow{9}{*}{\xmark} & ResNet-18 & 49.6 & 65.7 & - & - & 68.4 & 83.5 \\
         RelatNet \cite{relation_net}& & ResNet-18 & 52.5 & 69.8 &  - & - & 68.6 & 84.0 \\
         MatchNet \cite{matching_net}& & ResNet-18 & 52.9 & 68.9 &  - & - & 73.5 & 84.5 \\
         ProtoNet \cite{prototypical_nets}& & ResNet-18 & 54.2 & 73.4 &  - & - & 73.0 & 86.6 \\
         MTL \cite{sun2019meta} & & ResNet-12 & 61.2 & 75.5 & - & - &  - & - \\
         vFSL \cite{variational_fsl} & & ResNet-12 & 61.2 & 77.7 & - & - & - & - \\
         Neg-cosine \cite{liu2020negative}& & ResNet-18 & 62.3 & 80.9 & - & - & 72.7 & 89.4 \\
         MetaOpt \cite{lee2019meta} & & ResNet-12 & 62.6 & 78.6 & 66.0 & 81.6 & - & -\\
         SimpleShot \cite{simpleshot} & & ResNet-18 & 62.9 & 80.0 & 68.9 & 84.6 & 68.9 & 84.0 \\
         Distill \cite{tian2020rethinking} & & ResNet-12 & 64.8 & 82.1 & 71.5 & 86.0 & - & -\\
         \hline
         RelatNet + T \cite{can} & & ResNet-12 & 52.4 & 65.4 & - & - & - & -\\
         ProtoNet + T \cite{can} & & ResNet-12 & 55.2 & 71.1 & - & - & - & -\\
         MatchNet+T \cite{can} &  & ResNet-12 & 56.3 & 69.8 & - & - & - & -\\
         TPN \cite{liu2018learning} &  & ResNet-12 & 59.5 & 75.7 & - & - & - & -\\
         TEAM \cite{team}& & ResNet-18 & 60.1 & 75.9 & - & - & - & -\\
         Ent-min \cite{dhillon2019baseline}& & ResNet-12 & 62.4 & 74.5 & 68.4 & 83.4 & - & -\\
         CAN+T \cite{can}& & ResNet-12 & 67.2 & 80.6 & 73.2 & 84.9 & - & -\\
         LaplacianShot \cite{Laplacian}& & ResNet-18 & 72.1 & 82.3 & 79.0 & 86.4 & 81.0 & 88.7 \\
         \rowcolor{Gray} TIM-ADM & & ResNet-18 & 73.6 & \textbf{85.0} & \textbf{80.0} & \textbf{88.5} & 81.9 & 90.7 \\
         \rowcolor{Gray} TIM-GD & \multirow{-10}{*}{\cmark} & ResNet-18 & \textbf{73.9} & \textbf{85.0} & 79.9 & \textbf{88.5} & \textbf{82.2} & \textbf{90.8} \\
         \midrule
         LEO \cite{leo}& \multirow{5}{*}{\xmark} & WRN28-10 & 61.8 & 77.6 & 66.3 & 81.4 & - & -\\
         SimpleShot \cite{simpleshot}& & WRN28-10 & 63.5 & 80.3 & 69.8 & 85.3 & - & -\\
         MatchNet \cite{matching_net}& & WRN28-10 & 64.0 & 76.3 & - & - & - & -\\
         CC+rot+unlabeled \cite{gidaris2019boosting} & & WRN28-10 & 64.0 & 80.7 & 70.5 & 85.0 & - & -\\
         FEAT \cite{feat} & & WRN28-10 & 65.1 & 81.1 & 70.4 & 84.4 & - & -\\
         \hline
         AWGIM \cite{guo2020attentive}&  & WRN28-10 & 63.1 & 78.4 & 67.7 & 82.8 & - & -\\
         Ent-min \cite{dhillon2019baseline} &  & WRN28-10 & 65.7 & 78.4 & 73.3 & 85.5 & - & -\\
         SIB  \cite{hu2020empirical} &  & WRN28-10 & 70.0 & 79.2 & - & - & - & -\\
         BD-CSPN \cite{prototype}&  & WRN28-10 & 70.3 & 81.9 & 78.7 & 86.92 & - & -\\
         LaplacianShot  \cite{Laplacian} &  & WRN28-10 & 74.9 & 84.1 & 80.2 & 87.6 & - & -\\
         \rowcolor{Gray} TIM-ADM & & WRN28-10 & 77.5 & 87.2 & 82.0 & 89.7 & - & -\\
         \rowcolor{Gray} TIM-GD & \multirow{-7}{*}{\cmark} & WRN28-10 & \textbf{77.8} & \textbf{87.4} & \textbf{82.1} & \textbf{89.8} & - & -\\
         \bottomrule
    \end{tabular}
    \label{tab:benchmark_results}
    \end{table*}
    
    \subsection{Datasets}
        We provide an extensive evaluation of TIM the following few-shot learning benchmarks:\\
        
        \noindent\textbf{Standard benchmarks:} Standard benchmarks all use standard $K$-shot $N$-ways task generation procedures. Specifically, we experiment on:
        \begin{itemize}
            \item \textbf{\textit{mini}-Imagenet} benchmark introduced in \cite{matching_net} dataset contains with 100 classes, split as in \cite{ravi2016optimization} in 64/16/20 classes for training/validation/testing. Each class contains exactly 600 images. 
            \item \textbf{Caltech-UCSD Birds 200} (CUB) dataset \cite{cub} possesses 200 classes, split in 100/50/50 classes for training/validation/testing. Each class contains approximatively 60 images. 
            \item \textbf{\textit{Tiered}-Imagenet} \cite{tiered_imagenet} dataset is composed of 608 classes. The train/val/test split of classes is 351/97/160. Each class contains close to 1300 images. 
        \end{itemize}
        While these benchmarks have been traditionally used to evaluate few-shot learning methods, their fixed task format causes a problem as to the realism of the evaluation. In fact, \cite{cao2019theoretical} showed that using the same number of shots during training and evaluation already represents a learning bias. \\

        \noindent\textbf{\textsc{Meta-dataset}:} To complement our experiments on standard benchmarks, we use the recently introduced \textsc{Meta-dataset} \cite{triantafillou2019meta}. \textsc{Meta-dataset} aggregates the most popular image classification benchmarks. In total, it combines 10 different datasets including the well known ImageNet dataset. For each dataset, the classes are split between trainining/validation/testing, roughly following the 70\%/15\%/15\% proportion. For instance, ImageNet classes are split in 712/158/130 train/val/test classes. Therefore, the first challenge of \textsc{Meta-dataset} lies in the presence of domain shift between base training set and test set. Second, \textsc{Meta-dataset} offers a significantly more challenging task generation process than the standard $K$-ways $N$-shot tasks. In particular, each task has a random number of ways, support and query shots. Moreoever, the number of support samples varies across classes within a task. We refer the reader to \cite{triantafillou2019meta} for more details on the task generation process and the exact splits of each of the 10 datasets present in  \textsc{Meta-dataset}.
    
    \subsection{Hyperparameters} 
        \textbf{Standard benchmarks: }Hyperparameters for TIM are kept fixed across benchmark experiments for both methods TIM-GD and TIM-ADM. Specifically, the conditional entropy weight $\alpha$ and the cross-entropy weights $\lambda$ in Objective (\ref{eq:tim_objective}) are both set to $0.1$, and the penalty weight $\beta$ is set to 1. The temperature parameter $\tau$ in the classifier is set to 15. For TIM-GD method, we use the ADAM optimizer with the recommended parameters \cite{kingma2014adam}, and run 1000 iterations for each task. For TIM-ADM, we run 150 iterations. \\
        
        \noindent
        \textbf{\textsc{Meta-dataset}: } Following the procedure of \cite{triantafillou2019meta}, the hyperparameters of each method are tuned (following the instructions of each method) on the validation split of ImageNet ILSVRC 2012, both for TIM methods and reproduced methods.

    \subsection{Base-training procedure}
    
        \textbf{Standard benchmarks: } For \textit{mini}-ImageNet, \textit{tiered}-ImageNet and CUB, we following previous literature \cite{closer_look,Laplacian} and resort to two architectures ResNet-18 and WRN28-10 \cite{he2016deep} as feature extractors. We train the feature extractors with standard cross-entropy training on the base classes, with label smoothing. We emphasize that base training does not involve any meta-learning or episodic training strategy. The label-smoothing parameter is set to 0.1. The models are trained for 90 epochs, with the learning rate initialized to 0.1, and divided by 10 once halfway through (epochs 45) and once at 3/4 of the training (epoch 66). Batch size is set to 256 for ResNet-18, and to 128 for WRN28-10. During training, all the images are resized to $84 \times 84$, and we used the same data augmentation procedure as in \cite{Laplacian}, which includes random cropping, color jitter  and random horizontal flipping. \\
        
        \noindent
        \textbf{\textsc{Meta-dataset}: }For the newly introduced \textsc{Meta-dataset}, we reimplement the data  pipeline from scratch in PyTorch \footnote{We found the original Tensorflow implementation prohibitively slow when plugged in a Pytorch code. We make our reimplementation publicly available at \url{https://github.com/mboudiaf/pytorch-meta-dataset}} and we reproduce all compared methods in our framework. For non-episodic methods, we train a Resnet-18 for 100'000 iterations on the train split of ImageNet ILSVRC 2012 \cite{imagenet}. Except for the total number of iterations, we use the exact same training procedure as for standard benchmarks. We also reproduce the Proto-Net \cite{prototypical_nets} episodic baseline, for which we also train a Resnet-18 for 100'000 episodes, and same hyperparameters/augmentations as non-episodic methods. Note that contrary to \cite{triantafillou2019meta}, we train with fixed size episodes, as doing otherwise would represent an unfair learning bias for the episodic method (i.e knowing the testing task generation process prior to testing). Each training iteration represents  two 5-ways 5-shot and 20 query shots episodes, such that the number of samples processed for each batch (250) amounts the batch size of non-episodic methods (256).

    \subsection{Comparison on standard benchmarks}\label{sec:benchmark}

        We first evaluate our methods TIM-GD and TIM-ADM on the widely adopted \textit{mini}-ImageNet, \textit{tiered}-ImageNet and CUB benchmark datasets, in the most common 1-shot 5-way and 5-shot 5-way scenarios, with 15 query shots for each class. Results are reported in \autoref{tab:benchmark_results}, and are averaged over 10,000 episodes, following \cite{simpleshot}. We can observe that both TIM-GD and TIM-ADM yield state-of-the-art performances, consistently across all standard datasets, scenarios and backbones, improving over both transductive and inductive methods by significant margins. 

    \subsection{Beyond standard benchmarks} \label{sec:beyond_standard_benchmarks}
    
        \textbf{Impact of domain-shift: } Chen et al. \cite{closer_look} recently showed that the performance of most meta-learning methods may drop drastically when a domain-shift exists between the base training data and test data. Surprisingly, the simplest 
        discriminative baseline exhibited the best performance in this case. Therefore, we evaluate our methods in this challenging scenario. To this end, we simulate a domain shift by training the feature encoder on \textit{mini}-Imagenet while evaluating the methods on CUB, similarly to the setting introduced in \cite{closer_look}. TIM-GD and TIM-ADM beat previous methods by significant margins in the domain-shift scenario, consistently with our results in the standard few-shot benchmarks, thereby demonstrating an increased potential of applicability to real-world situations. \\

        \begin{table}
            \newlength\wexp
            \settowidth{\wexp}{\textbf{\textit{mini}-ImageNet} $\rightarrow$ \textbf{CUB}}
            \centering
            \small
            \caption{The results for the domain-shift setting \textit{mini}-Imagenet $\rightarrow$ CUB. All methods use a ResNet-18 as backbone. The results obtained by our models (gray-shaded) are averaged over 10,000 episodes.}
            \begin{tabular}{lcccc}
                \toprule
                & \multicolumn{1}{c}{\textbf{\textit{mini}-ImageNet} $\rightarrow$ \textbf{CUB}}\\
                 Methods & 5-shot\\
                 \midrule
                 MatchNet \cite{matching_net} &  53.1\\
                 MAML \cite{maml} & 51.3 \\
                 ProtoNet \cite{prototypical_nets} & 62.0 \\
                 RelatNet \cite{relation_net} & 57.7 \\
                SimpleShot \cite{simpleshot} & 64.0 \\
                GNN \cite{tseng2020cross} & 66.9 \\
                Neg-Cosine \cite{liu2020negative} & 67.0 \\
                Baseline \cite{closer_look} & 65.6 \\
                LaplacianShot \cite{Laplacian} & 66.3 \\
                \rowcolor{Gray} TIM-ADM & 70.3 \\
                \rowcolor{Gray} TIM-GD & \textbf{71.0} \\
                 \bottomrule
            \end{tabular}
            \label{tab:domin_shift_results}
        \end{table}
         
         \noindent
        \textbf{Increasing the number of ways: } Most few-shot papers only evaluate their method in the usual 5-ways scenario. Nevertheless, \cite{closer_look} showed that meta-learning methods could be beaten by their discriminative baseline when more ways were introduced in each task. Therefore, we also provide results of our method in the more challenging 10-ways and 20-ways scenarios on \textit{mini}-ImageNet. These results, which are presented in \autoref{tab:more_ways}, show that TIM-GD outperforms other methods by significant margins, in both settings. \\
        
        \begin{table}
            \centering
            \caption{Results for increasing the number of classes on \textit{mini}-ImageNet. The results obtained by our models (gray-shaded) are averaged over 10,000 episodes.}
            \small
            \begin{tabular}{lcccccccc}
                \toprule
                & \multicolumn{2}{c}{10-way} & \multicolumn{2}{c}{20-way} \\
                 \cmidrule(lr){2-3}\cmidrule(lr){4-5}
                 Methods & 1-shot & 5-shot & 1-shot & 5-shot\\
                 \midrule
                 MatchNet \cite{matching_net}& - & 52.3 & - & 36.8 \\
                 ProtoNet \cite{prototypical_nets}&  - & 59.2 & - & 45.0 \\
                 RelatNet \cite{relation_net}& - & 53.9 & - & 39.2 \\
                 SimpleShot \cite{simpleshot}& 45.1 & 68.1 & 32.4 & 55.4 \\
                 Baseline \cite{closer_look}& - & 55.0 & - & 42.0 \\
                 Baseline++ \cite{closer_look}& - & 63.4 & - & 50.9 \\
                 \rowcolor{Gray} TIM-ADM & 56.0 & \textbf{72.9} & \textbf{39.5} & 58.8 \\
                 \rowcolor{Gray} TIM-GD & \textbf{56.1} & 72.8 & 39.3 & \textbf{59.5} \\
                 \bottomrule
            \end{tabular}
            \label{tab:more_ways}
        \end{table}

        \noindent
        \textbf{Random tasks and domain shift: } More recently, the \textsc{Meta-dataset} \cite{triantafillou2019meta} was introduced to provide a more realistic evaluation of few-shot methods. \textsc{Meta-dataset} combines both randomness of number of samples and of ways, as well as domain-shift scnearios. To first validate our PyTorch implementaiton, we provide a comparison between the performances of the SimpleShot \cite{simpleshot} baseline obtained with the original implementation and our implementation in \autoref{tab:meta_dataset_validation}. \textbf{We found a significant difference of 3 \% on average, that we eventually identified to be due to the absence of Anti-aliasing when resizing images in the original implementation of \cite{triantafillou2019meta}}. More details on this can be found in the supplementary material. To provide the fairest comparison possible, we reproduce all methods with our implementation. The results are provided in \autoref{tab:meta_dataset_results}. TIM-GD appears as the best overall performing method, followed by TIM-ADM. The simple inductive \textit{Finetune} baseline achieves impressive performance, even above the transductive method BD-CSPN \cite{prototype}. Note that the episodic ProtoNet baseline performs dramatically worse than other inductive baselines, which we hypothesize is due to the fact it was trained on fixed-size episodes, but tested on random tasks.
        
        \begin{table*}
            \centering
            \small
            \caption{Validation of our Pytorch implementation of \textsc{Meta-dataset} \cite{triantafillou2019meta}. We report the results of SimpleShot \cite{simpleshot} with a Resnet-18. We found activating anti-aliasing when resizing images to 84x84 in the original code of \cite{triantafillou2019meta} (deactivated by default) can significantly improve the performances. More details can be found in the supplementary material. Results are averaged over 600 random episodes.}
            \resizebox{\textwidth}{!}{
                \begin{tabular}{lccccccccccc}
                    \toprule
                     Implementation & ILSVRC & Omniglot & Aircraft & Birds & Textures & Quick Draw & Fungi & VGG Flower & Traffic Signs & MSCOCO & Mean\\
                     \midrule
                     Original & 52.7 & 53.9 & 50.1 & 68.5 & 74.6 & 57.5 & 42.7 & 87.5 & 49.7 & 44.3 & 58.1 \\                    
                     Original + anti-aliasing & 59.7 & 54.0 & 55.1 & 79.1 & 77.4 & 56.9 & 48.7 & 90.8 & 49.3 & 45.9 & 61.7 \\
                     Ours & 60.0 & 54.2 & 55.9 & 78.6 & 77.8 & 57.4 & 49.2 & 90.3 & 49.6 & 44.2 & 61.7 \\
                     \bottomrule
                \end{tabular}
            }
            \label{tab:meta_dataset_validation}
        \end{table*}

        \begin{table*}
            \centering
            \small
            \caption{Comparison to the state-of-the-art methods on \textsc{Meta-dataset}. All methods use a Resnet-18 backbone trained on ILSVRC 2012 only, and tested on the test split of each of the following datasets, using our Pytorch implementation of \textsc{Meta-dataset} \cite{triantafillou2019meta}. Methods are grouped in inductive (Ind.) and transductive (Transd.) group. Following \cite{triantafillou2019meta}, results are averaged over 600 random episodes.}
            \resizebox{\textwidth}{!}{
                \begin{tabular}{llcccccccccccc}
                    \toprule
                     & Method & ILSVRC & Omniglot & Aircraft & Birds & Textures & Quick Draw & Fungi & VGG Flower & Traffic Signs & MSCOCO & Mean \\
                     \midrule
                     \multirow{3}{*}{\begin{turn}{90} Ind. \end{turn}} 
                     & Finetune \cite{closer_look} & 59.8 & 60.5 & 63.5 & 80.6 & 80.9 & 61.5 & 45.2 & 91.1 & 55.1 & 41.8 & 64.0 \\
                     & ProtoNet \cite{prototypical_nets} & 48.2 & 46.7 & 44.6 & 53.8 & 70.3 & 45.1 & 38.5 & 82.4 & 42.2 & 38.0 & 51.0  \\
                     & SimpleShot \cite{simpleshot} & 60.0 & 54.2 & 55.9 & 78.6 & 77.8 & 57.4 & 49.2 & 90.3 & 49.6 & 44.2 & 61.7 \\
                     \midrule
                     & BD-CSPN \cite{prototype} & 60.5 & 54.4 & 55.2 & 80.9 & 77.9 & 57.3 & 50.0 & 91.7 & 47.8 & 43.9 & 62.0 \\
                     \rowcolor{Gray} & TIM-ADM & \textbf{64.1} & 62.4 & 63.8 & 84.6 & 84.1 & 63.4 & \textbf{57.6} & 94.4 & 62.9 & \textbf{52.1} & 68.9  \\
                     \rowcolor{Gray} \multirow{-3}{*}{\begin{turn}{90} Transd. \end{turn}}   & TIM-GD & 63.6 & \textbf{65.6} & \textbf{66.4} & \textbf{85.6} & \textbf{84.7} & \textbf{65.8} & 57.5 & \textbf{95.6} & \textbf{65.2} & 50.9 & \textbf{70.1} \\
                     \bottomrule
                \end{tabular}
            }
            \label{tab:meta_dataset_results}
        \end{table*}
        
        \begin{table*}[t]
           \small
           \centering
            \caption{Ablation study on the effect of each term in our loss in Eq. \eqref{eq:tim_objective}, when only the classifier weights are fine-tuned, i.e., updating only $\textbf{W}$, and when the whole network is fine-tuned, i.e., updating $\{\boldsymbol{\phi},\textbf{W}\}$. The results are reported for ResNet-18 as backbone. 
            The same term indexing as in Eq. \eqref{eq:tim_objective} is used here: $\mathcal{H}(\hat{Y}_\mathcal{Q})$: Marginal entropy, $\mathcal{H}(\hat{Y}_\mathcal{Q}|X_\mathcal{Q})$: Conditional entropy, $\mathrm{CE}$: Cross-entropy.}
           \resizebox{0.8\textwidth}{!}{
           \begin{tabular}{ccccccccc}
                \toprule
                 & & & \multicolumn{2}{c}{\textbf{\textit{mini}-ImageNet}} & \multicolumn{2}{c}{\textbf{\textit{tiered}-ImageNet}} & \multicolumn{2}{c}{\textbf{CUB}}\\
                 \cmidrule(lr){4-5}\cmidrule(lr){6-7}\cmidrule(lr){8-9}
                 Method & Param. & Loss & 1-shot & 5-shot & 1-shot & 5-shot & 1-shot & 5-shot \\
                 \midrule
                 \multirow{4}{*}{TIM-ADM} & \multirow{4}{*}{$\{\textbf{W}\}$} & $\mathrm{CE}$ & 60.0 & 79.6 & 68.0 & 84.6 & 68.6 & 86.4 \\
                 & & $\mathrm{CE} + \mathcal{H}(\hat{Y}_\mathcal{Q}|X_\mathcal{Q})$ & 36.0 & 77.0 & 48.1 & 82.5 & 48.5 & 86.5 \\
                 & & $\mathrm{CE} - \mathcal{H}(\hat{Y}_\mathcal{Q})$ & 66.7 & 82.0 & 74.0 & 86.5 & 74.2 & 88.3 \\
                 & & $\mathrm{CE} - \mathcal{H}(\hat{Y}_\mathcal{Q}) + \mathcal{H}(\hat{Y}_\mathcal{Q}|X_\mathcal{Q})$ & \textbf{73.6} & \textbf{85.0} & \textbf{80.0} & \textbf{88.5} & \textbf{81.9} & \textbf{90.7} \\
                 \hline
                 \multirow{4}{*}{TIM-GD} & \multirow{4}{*}{\{$\textbf{W} $\}}& $\mathrm{CE}$ & 60.7 & 79.4 & 68.4 & 84.3 & 69.6 & 86.3 \\
                 & & $\mathrm{CE} + \mathcal{H}(\hat{Y}_\mathcal{Q}|X_\mathcal{Q})$ & 35.3 & 79.2 & 45.9 & 80.6 & 46.1 & 85.9 \\
                 & & $\mathrm{CE} - \mathcal{H}(\hat{Y}_\mathcal{Q})$ & 66.1 & 81.3 & 73.4 & 86.0 & 73.9 & 88.0 \\
                 & & $\mathrm{CE} - \mathcal{H}(\hat{Y}_\mathcal{Q}) + \mathcal{H}(\hat{Y}_\mathcal{Q}|X_\mathcal{Q})$ & \textbf{73.9} & \textbf{85.0} & \textbf{79.9} & \textbf{88.5} & \textbf{82.2} & \textbf{90.8} \\
                 \hline
                 \multirow{4}{*}{TIM-GD} & \multirow{4}{*}{$\{\boldsymbol{\phi}, \textbf{W}\}$}& $\mathrm{CE}$ & 60.8 & 81.6 & 65.7 & 83.5 & 68.7 & 87.7\\
                 & & $\mathrm{CE} + \mathcal{H}(\hat{Y}_\mathcal{Q}|X_\mathcal{Q})$ & 62.7 & 81.9 & 66.9 & 82.8 & 72.6 & 89.0 \\
                 & & $\mathrm{CE} - \mathcal{H}(\hat{Y}_\mathcal{Q})$ & 62.3 & 82.7 & 68.3 & 85.4 & 70.7 & 88.8 \\
                 & & $\mathrm{CE} - \mathcal{H}(\hat{Y}_\mathcal{Q}) + \mathcal{H}(\hat{Y}_\mathcal{Q}|X_\mathcal{Q})$ & \textbf{67.2} & \textbf{84.7} & \textbf{73.0} & \textbf{86.8} & \textbf{76.7} & \textbf{90.5} \\
                 \bottomrule
            \end{tabular}}
            \label{tab:ablation_effect_terms}
        \end{table*}
            
    \subsection{Ablation study}

        \textbf{Influence of each term: }\label{sec:ablation_terms} We now assess the impact of each term\footnote{The $\textbf{W}$ and $\textbf{q}$ updates of TIM-ADM associated to each configuration can be found in the supplementary material.} in our loss in Eq. (\ref{eq:tim_objective}) on the final performance of our methods. The results are reported in \autoref{tab:ablation_effect_terms}. 
        We observe that integrating the three terms in our loss consistently outperforms any other configuration. Interestingly, removing the label-marginal entropy, $\mathcal{H}(\hat{Y}_\mathcal{Q})$, reduces significantly the performances in both TIM-GD and TIM-ADM, particularly when only classifier weights $\textbf{W}$ are 
        updated and feature extractor $\boldsymbol{\phi}$ is fixed. 
        Such a behavior could be explained by the following fact: the conditional entropy term, $\mathcal{\widehat{H}}(Y_\mathcal{Q}|X_\mathcal{Q})$, may yield degenerate solutions (assigning all query samples to a single class) on numerous tasks, when used alone. 
        This emphasizes the importance of the label-marginal entropy term $\mathcal{H}(\hat{Y}_\mathcal{Q})$ in our loss (\ref{eq:tim_objective}), which acts as a powerful regularizer to prevent such trivial solutions. \\

\begin{figure*}[t]
            \centering
            \includegraphics[width=0.7\textwidth]{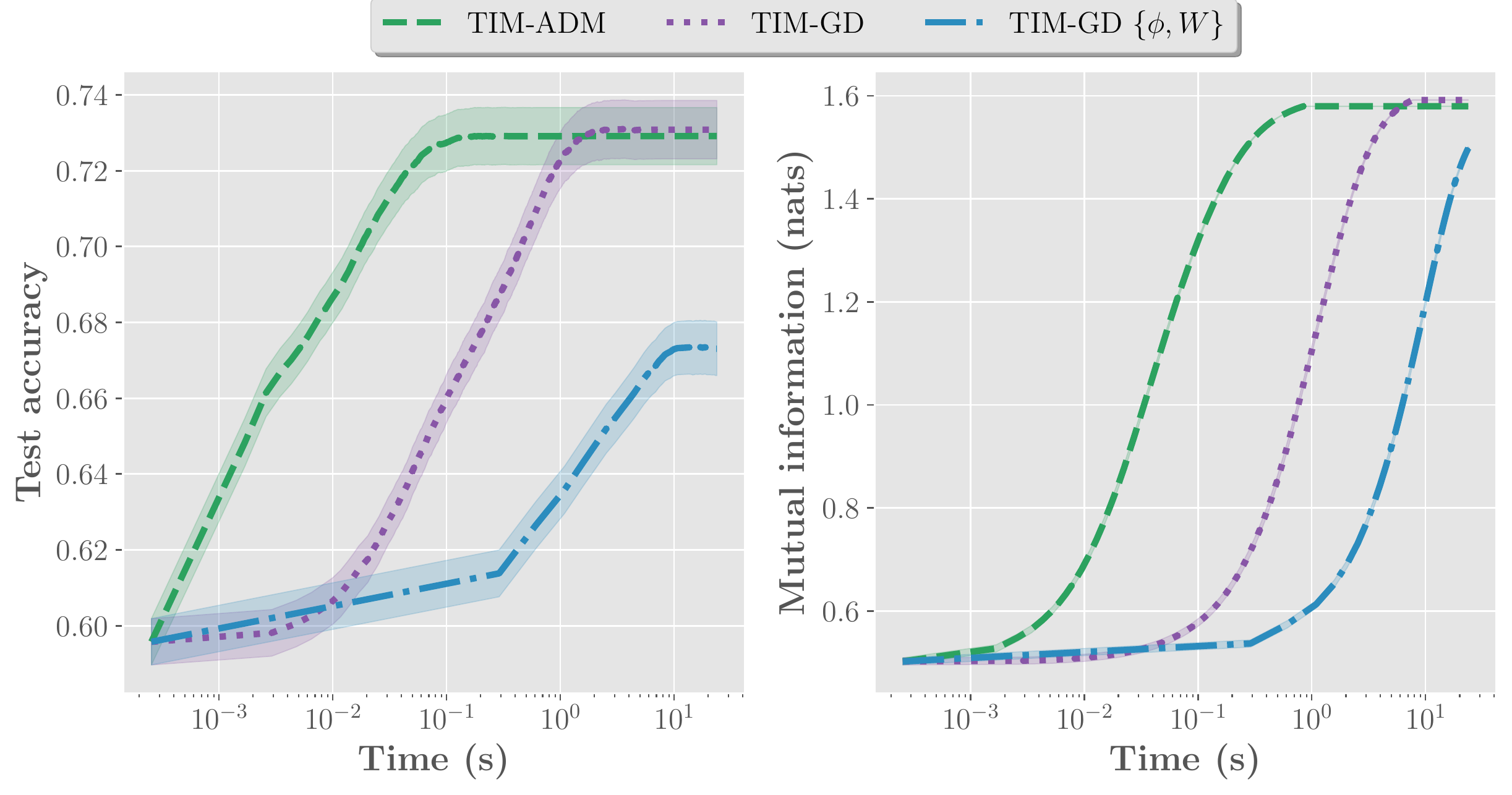}
            \caption{Convergence plots for our methods on \textit{mini}-ImageNet with a ResNet-18. Solid lines are averages, while shadows are 95\% confidence intervals. Time is in logarithmic scale. \textbf{Left:} Evolution of the test accuracy during transductive inference. \textbf{Right:} Evolution of the mutual information between query features and predictions $\widehat{\mathcal{I}}(X_\mathcal{Q};Y_\mathcal{Q})$, computed as in Eq. (\ref{prior-aware-MI}), with $\alpha=1$.}
            \label{fig:convergence_methods}
        \end{figure*}

        \noindent
        \textbf{Fine-tuning the whole network vs classifier only: }
            While our TIM-GD and TIM-ADM optimize w.r.t $\textbf{W}$ and keep base-trained encoder $f_{\boldsymbol{\phi}}$ fixed at inference, the authors of \cite{dhillon2019baseline} fine-tuned the whole network $\{\textbf{W}, \boldsymbol{\phi}\}$ when performing their transductive entropy minimization. To assess both approaches, we add to \autoref{tab:ablation_effect_terms} a variant of TIM-GD, in which we fine-tune the whole network $\{\textbf{W}, \boldsymbol{\phi}\}$, by using the same optimization procedure as in \cite{dhillon2019baseline}. We found that, besides being much slower, fine-tuning the whole network for our objective in Eq. \ref{eq:tim_objective} degrades the performances, as also conveyed by the convergence plots in \autoref{fig:convergence_methods}. 
            Interestingly, when fine-tuning the whole network $\{\textbf{W}, \boldsymbol{\phi}\}$,  the absence of $\mathcal{H}(\hat{Y}_\mathcal{Q})$ in the entropy-based loss $\mathrm{CE} + \mathcal{H}(\hat{Y}_\mathcal{Q}|X_\mathcal{Q})$ does not cause the same drastic drop in performance as observed earlier when optimizing with respect to $\textbf{W}$ only. We hypothesize that the network's intrinsic regularization (such as batch normalizations) and the use of small learning rates, as prescribed by \cite{dhillon2019baseline}, help the optimization process, preventing the predictions from approaching the vertices of the simplex, where entropy's gradients diverge.

    \subsection{Inference run-times}
    
        Transductive methods are generally slower at inference than their inductive counterparts, with run-times that are, typically, several orders of magnitude larger. In \autoref{tab:runtimes}, we measure the average adaptation time per few-shot task, defined as the time required by each method to build the final classifier, for a 5-shot 5-way task on \textit{mini}-ImageNet using the WRN28-10 network. Table \ref{tab:runtimes} conveys that our ADM optimization gains one order of magnitude in run-time over our gradient-based method, and more than two orders of magnitude in comparison to \cite{dhillon2019baseline}, which fine-tunes the whole network. Note that TIM-ADM still remains slower than the inductive baseline. Our methods were run on the same GTX 1080 Ti GPU, while the run-time of \cite{dhillon2019baseline} is directly reported from the paper. \\

        \begin{table}[t]
            \centering
            \resizebox{\textwidth}{!}{
                \begin{tabular}{lccc}
                    \toprule
                    Method & Parameters & Transductive & Inference/task (s) \\
                    \toprule
                    SimpleShot \cite{simpleshot}& $\{\textbf{W}\}$ & \xmark & $9.0 \times 10^{-3}$ \\
                    \midrule
                    \rowcolor{Gray} TIM-ADM & $\{\textbf{W}\}$ & & $1.2 \times 10^{-1}$\\
                    \rowcolor{Gray} TIM-GD & $\{\textbf{W}\}$ & & $2.2 \times 10^{+0}$\\
                    Ent-min \cite{dhillon2019baseline} & $\{\boldsymbol{\phi}, \textbf{W}\}$ & \multirow{-3}{*}{\cmark} & $2.1 \times 10^{+1}$ \\
                    \bottomrule
                \end{tabular}
            }
            \caption{Inference run-time per few-shot task for a 5-shot 5-way task on mini-ImageNet with a WRN28-10 backbone.}
            \label{tab:runtimes}
        \end{table}

\section{Conclusion}

    TIM inference establishes new state-of-the-art results on the standard few-shot benchmarks, as well as in more challenging scenarios, with random numbers of classes, of samples and domain shifts. We used feature extractors based on a simple base-class training with the standard cross-entropy loss, without resorting to the complex meta-training schemes that are often used and advocated in the recent few-shot literature. TIM is modular: it could be plugged on top of any feature extractor and base training, regardless of how the training was conducted. Therefore, while we do not claim that the very challenging few-shot problem is solved, we believe that our model-agnostic TIM inference should be used as a strong baseline for future few-shot learning research. 

\section*{Acknowledgements}
We thank Hoel Kervadec for insightful comments, noticing the absence of antialiasing in the original implementation of \cite{triantafillou2019meta} and thoroughly reviewing the code. This research was supported by the National Science and Engineering Research Council of Canada (NSERC), via its Discovery Grant program. The work of Prof. Pablo Piantanida was supported by the European Commission’s Marie Sklodowska-Curie Actions (MSCA), through the Marie Sklodowska-Curie IF (H2020-MSCAIF-2017-EF-797805).

\bibliographystyle{abbrv}
\bibliography{main}

\begin{thebibliography}{10}

\bibitem{Barlow1989Unsupervised}
H.~B. Barlow.
\newblock Unsupervised learning.
\newblock In {\em Neural Comput.}, 1989.

\bibitem{berthelot2019mixmatch}
D.~Berthelot, N.~Carlini, I.~Goodfellow, N.~Papernot, A.~Oliver, and C.~A.
  Raffel.
\newblock Mixmatch: A holistic approach to semi-supervised learning.
\newblock In {\em Advances in Neural Information Processing Systems (NeurIPS)},
  2019.

\bibitem{boudiaf2020transductive}
M.~Boudiaf, Z.~I. Masud, J.~Rony, J.~Dolz, P.~Piantanida, and I.~B. Ayed.
\newblock Transductive information maximization for few-shot learning.
\newblock {\em Advances in Neural Information Processing Systems (NeurIPS)},
  2020.

\bibitem{boudiaf2020metric}
M.~Boudiaf, J.~Rony, I.~M. Ziko, E.~Granger, M.~Pedersoli, P.~Piantanida, and
  I.~B. Ayed.
\newblock A unifying mutual information view of metric learning: cross-entropy
  vs. pairwise losses.
\newblock In {\em European Conference on Computer Vision (ECCV)}, 2020.

\bibitem{boyd}
S.~Boyd, N.~Parikh, E.~Chu, B.~Peleato, and J.~Eckstein.
\newblock Distributed optimization and statistical learning via the alternating
  direction method of multipliers.
\newblock In {\em Foundations and Trends{\textregistered} in Machine learning}.
  Now Publishers Inc., 2011.

\bibitem{cao2019theoretical}
T.~Cao, M.~Law, and S.~Fidler.
\newblock A theoretical analysis of the number of shots in few-shot learning.
\newblock {\em International Conference on Learning Representations (ICLR)},
  2020.

\bibitem{closer_look}
W.-Y. Chen, Y.-C. Liu, Z.~Kira, Y.-C.~F. Wang, and J.-B. Huang.
\newblock A closer look at few-shot classification.
\newblock In {\em International Conference on Learning Representations (ICLR)},
  2019.

\bibitem{csiszar2011}
I.~Csisz{\'a}r and J.~K{\"o}rner.
\newblock Information theory: Coding theorems for discrete memoryless systems.
\newblock 2011.

\bibitem{z2004learning}
Z.~Dengyong, O.~Bousquet, T.~N. Lal, J.~Weston, and B.~Sch{\"o}lkopf.
\newblock Learning with local and global consistency.
\newblock In {\em Advances in Neural Information Processing Systems (NeurIPS)},
  2004.

\bibitem{dhillon2019baseline}
G.~S. Dhillon, P.~Chaudhari, A.~Ravichandran, and S.~Soatto.
\newblock A baseline for few-shot image classification.
\newblock In {\em International Conference on Learning Representations (ICLR)},
  2020.

\bibitem{fei2006one}
L.~Fei-Fei, R.~Fergus, and P.~Perona.
\newblock One-shot learning of object categories.
\newblock In {\em IEEE Transactions on Pattern Analysis and Machine
  Intelligence (TPAMI)}, 2006.

\bibitem{maml}
C.~Finn, P.~Abbeel, and S.~Levine.
\newblock Model-agnostic meta-learning for fast adaptation of deep networks.
\newblock In {\em International Conference on Machine Learning (ICML)}, 2017.

\bibitem{gidaris2019boosting}
S.~Gidaris, A.~Bursuc, N.~Komodakis, P.~P{\'e}rez, and M.~Cord.
\newblock Boosting few-shot visual learning with self-supervision.
\newblock In {\em International Conference on Computer Vision (ICCV)}, 2019.

\bibitem{grandvalet2005semi}
Y.~Grandvalet and Y.~Bengio.
\newblock Semi-supervised learning by entropy minimization.
\newblock In {\em Advances in neural information processing systems (NeurIPS)},
  2005.

\bibitem{gunawardana2005convergence}
A.~Gunawardana, W.~Byrne, and M.~I. Jordan.
\newblock Convergence theorems for generalized alternating minimization
  procedures.
\newblock {\em Journal of machine learning research}, 6(12), 2005.

\bibitem{guo2020attentive}
Y.~Guo and N.-M. Cheung.
\newblock Attentive weights generation for few shot learning via information
  maximization.
\newblock In {\em IEEE Conference on Computer Vision and Pattern Recognition
  (CVPR)}, 2020.

\bibitem{he2016deep}
K.~He, X.~Zhang, S.~Ren, and J.~Sun.
\newblock Deep residual learning for image recognition.
\newblock In {\em Proceedings of the IEEE conference on computer vision and
  pattern recognition}, pages 770--778, 2016.

\bibitem{deep_infomax}
R.~D. Hjelm, A.~Fedorov, S.~Lavoie-Marchildon, K.~Grewal, P.~Bachman,
  A.~Trischler, and Y.~Bengio.
\newblock Learning deep representations by mutual information estimation and
  maximization.
\newblock In {\em International Conference on Learning Representations (ICLR)},
  2019.

\bibitem{can}
R.~Hou, H.~Chang, M.~Bingpeng, S.~Shan, and X.~Chen.
\newblock Cross attention network for few-shot classification.
\newblock In {\em Advances in Neural Information Processing Systems (NeurIPS)},
  2019.

\bibitem{hu2020empirical}
S.~X. Hu, P.~G. Moreno, Y.~Xiao, X.~Shen, G.~Obozinski, N.~D. Lawrence, and
  A.~Damianou.
\newblock Empirical bayes transductive meta-learning with synthetic gradients.
\newblock In {\em International Conference on Learning Representations (ICLR)},
  2020.

\bibitem{HuICML17}
W.~Hu, T.~Miyato, S.~Tokui, E.~Matsumoto, and M.~Sugiyama.
\newblock Learning discrete representations via information maximizing
  self-augmented training.
\newblock In {\em International Conference on Machine Learning (ICML)}, 2017.

\bibitem{jabi}
M.~Jabi, M.~Pedersoli, A.~Mitiche, and I.~B. Ayed.
\newblock Deep clustering: On the link between discriminative models and
  k-means.
\newblock In {\em IEEE Transactions on Pattern Analysis and Machine
  Intelligence (TPAMI)}, 2020.

\bibitem{joachim99}
T.~Joachims.
\newblock Transductive inference for text classification using support vector
  machines.
\newblock In {\em International Conference on Machine Learning (ICML)}, 1999.

\bibitem{kim2019edge}
J.~Kim, T.~Kim, S.~Kim, and C.~D. Yoo.
\newblock Edge-labeling graph neural network for few-shot learning.
\newblock In {\em IEEE Conference on Computer Vision and Pattern Recognition
  (CVPR)}, 2019.

\bibitem{kingma2014adam}
D.~P. Kingma and J.~Ba.
\newblock Adam: A method for stochastic optimization.
\newblock In {\em International Conference on Learning Representations (ICLR)},
  2014.

\bibitem{kovalevsky1968}
V.~A. Kovalevsky.
\newblock {The problem of character recognition from the point of view of
  mathematical statistics}.
\newblock In {\em Character Readers and Pattern Recognition}, New York, 1968.

\bibitem{clustering_infomax}
A.~Krause, P.~Perona, and R.~G. Gomes.
\newblock Discriminative clustering by regularized information maximization.
\newblock In {\em Advances in Neural Information Processing systems (NeurIPS)},
  2010.

\bibitem{lee2019meta}
K.~Lee, S.~Maji, A.~Ravichandran, and S.~Soatto.
\newblock Meta-learning with differentiable convex optimization.
\newblock In {\em IEEE Conference on Computer Vision and Pattern Recognition
  (CVPR)}, 2019.

\bibitem{paper_to_please_R1}
X.~Li, Q.~Sun, Y.~Liu, Q.~Zhou, S.~Zheng, T.-S. Chua, and B.~Schiele.
\newblock Learning to self-train for semi-supervised few-shot classification.
\newblock In {\em Advances in Neural Information Processing Systems (NeurIPS)},
  2019.

\bibitem{liang2020we}
J.~Liang, D.~Hu, and J.~Feng.
\newblock Do we really need to access the source data? source hypothesis
  transfer for unsupervised domain adaptation.
\newblock In {\em International Conference on Machine Learning (ICML)}, 2020.

\bibitem{Linsker1988Self}
R.~Linsker.
\newblock Self-organization in a perceptual network.
\newblock In {\em Computer}, 1988.

\bibitem{liu2020negative}
B.~Liu, Y.~Cao, Y.~Lin, Q.~Li, Z.~Zhang, M.~Long, and H.~Hu.
\newblock Negative margin matters: Understanding margin in few-shot
  classification.
\newblock In {\em European Conference on Computer Vision (ECCV)}, 2020.

\bibitem{prototype}
J.~Liu, L.~Song, and Y.~Qin.
\newblock Prototype rectification for few-shot learning.
\newblock In {\em European Conference on Computer Vision (ECCV)}, 2020.

\bibitem{liu2018learning}
Y.~Liu, J.~Lee, M.~Park, S.~Kim, E.~Yang, S.~J. Hwang, and Y.~Yang.
\newblock Learning to propagate labels: Transductive propagation network for
  few-shot learning.
\newblock In {\em International Conference on Learning Representations (ICLR)},
  2019.

\bibitem{miller2000learning}
E.~G. Miller, N.~E. Matsakis, and P.~A. Viola.
\newblock Learning from one example through shared densities on transforms.
\newblock In {\em IEEE Conference on Computer Vision and Pattern Recognition
  (CVPR)}, 2000.

\bibitem{Mishra18}
N.~Mishra, M.~Rohaninejad, X.~Chen, and P.~A. Abbeel.
\newblock simple neural attentive meta-learner.
\newblock In {\em International Conference on Learning Representations (ICLR)},
  2018.

\bibitem{miyato2018virtual}
T.~Miyato, S.-i. Maeda, M.~Koyama, and S.~Ishii.
\newblock Virtual adversarial training: a regularization method for supervised
  and semi-supervised learning.
\newblock In {\em IEEE Transactions on Pattern Analysis and Machine
  Intelligence (TPAMI)}, 2018.

\bibitem{nichol2018firstorder}
A.~Nichol, J.~Achiam, and J.~Schulman.
\newblock On first-order meta-learning algorithms.
\newblock In {\em arXiv preprint arXiv:1803.02999}, 2018.

\bibitem{tadam}
B.~Oreshkin, P.~R. L{\'o}pez, and A.~Lacoste.
\newblock Tadam: Task dependent adaptive metric for improved few-shot learning.
\newblock In {\em Advances in Neural Information Processing Systems (NeurIPS)},
  2018.

\bibitem{team}
L.~Qiao, Y.~Shi, J.~Li, Y.~Wang, T.~Huang, and Y.~Tian.
\newblock Transductive episodic-wise adaptive metric for few-shot learning.
\newblock In {\em International Conference on Computer Vision (ICCV)}, 2019.

\bibitem{ravi2016optimization}
S.~Ravi and H.~Larochelle.
\newblock Optimization as a model for few-shot learning.
\newblock In {\em International Conference on Learning Representations (ICLR)},
  2016.

\bibitem{tiered_imagenet}
M.~Ren, E.~Triantafillou, S.~Ravi, J.~Snell, K.~Swersky, J.~B. Tenenbaum,
  H.~Larochelle, and R.~S. Zemel.
\newblock Meta-learning for semi-supervised few-shot classification.
\newblock In {\em International Conference on Learning Representations (ICLR)},
  2018.

\bibitem{imagenet}
O.~Russakovsky, J.~Deng, H.~Su, J.~Krause, S.~Satheesh, S.~Ma, Z.~Huang,
  A.~Karpathy, A.~Khosla, M.~Bernstein, A.~C. Berg, and L.~Fei-Fei.
\newblock {ImageNet Large Scale Visual Recognition Challenge}.
\newblock In {\em International booktitle of Computer Vision (IJCV)}, 2015.

\bibitem{leo}
A.~A. Rusu, D.~Rao, J.~Sygnowski, O.~Vinyals, R.~Pascanu, S.~Osindero, and
  R.~Hadsell.
\newblock Meta-learning with latent embedding optimization.
\newblock In {\em International Conference on Learning Representations (ICLR)},
  2019.

\bibitem{prototypical_nets}
J.~Snell, K.~Swersky, and R.~Zemel.
\newblock Prototypical networks for few-shot learning.
\newblock In {\em Advances in neural information processing systems (NeurIPS)},
  2017.

\bibitem{sriperumbudur2009convergence}
B.~K. Sriperumbudur and G.~R. Lanckriet.
\newblock On the convergence of the concave-convex procedure.
\newblock In {\em Nips}, volume~9, pages 1759--1767. Citeseer, 2009.

\bibitem{sun2019meta}
Q.~Sun, Y.~Liu, T.-S. Chua, and B.~Schiele.
\newblock Meta-transfer learning for few-shot learning.
\newblock In {\em IEEE Conference on Computer Vision and Pattern Recognition
  (CVPR)}, 2019.

\bibitem{relation_net}
F.~Sung, Y.~Yang, L.~Zhang, T.~Xiang, P.~H. Torr, and T.~M. Hospedales.
\newblock Learning to compare: Relation network for few-shot learning.
\newblock In {\em IEEE Conference on Computer Vision and Pattern Recognition
  (CVPR)}, 2018.

\bibitem{tian2020rethinking}
Y.~Tian, Y.~Wang, D.~Krishnan, J.~B. Tenenbaum, and P.~Isola.
\newblock Rethinking few-shot image classification: a good embedding is all you
  need?
\newblock In {\em European Conference on Computer Vision (ECCV)}, 2020.

\bibitem{triantafillou2019meta}
E.~Triantafillou, T.~Zhu, V.~Dumoulin, P.~Lamblin, U.~Evci, K.~Xu, R.~Goroshin,
  C.~Gelada, K.~Swersky, P.-A. Manzagol, et~al.
\newblock Meta-dataset: A dataset of datasets for learning to learn from few
  examples.
\newblock {\em International Conference on Learning Representations (ICLR)},
  2020.

\bibitem{tseng2020cross}
H.-Y. Tseng, H.-Y. Lee, J.-B. Huang, and M.-H. Yang.
\newblock Cross-domain few-shot classification via learned feature-wise
  transformation.
\newblock In {\em International Conference on Learning Representations (ICLR)},
  2020.

\bibitem{cpc}
A.~van~den Oord, Y.~Li, and O.~Vinyals.
\newblock Representation learning with contrastive predictive coding.
\newblock In {\em ArXiv preprint arXiv:1807.03748}, 2019.

\bibitem{vapnik1999overview}
V.~N. Vapnik.
\newblock An overview of statistical learning theory.
\newblock In {\em IEEE Transactions on Neural Networks (TNN)}, 1999.

\bibitem{matching_net}
O.~Vinyals, C.~Blundell, T.~Lillicrap, D.~Wierstra, et~al.
\newblock Matching networks for one shot learning.
\newblock In {\em Advances in Neural Information Processing Systems (NeurIPS)},
  2016.

\bibitem{simpleshot}
Y.~Wang, W.-L. Chao, K.~Q. Weinberger, and L.~van~der Maaten.
\newblock Simpleshot: Revisiting nearest-neighbor classification for few-shot
  learning.
\newblock In {\em arXiv preprint arXiv:1911.04623}, 2019.

\bibitem{cub}
P.~Welinder, S.~Branson, T.~Mita, C.~Wah, F.~Schroff, S.~Belongie, and
  P.~Perona.
\newblock {Caltech-UCSD Birds 200}.
\newblock Technical Report CNS-TR-2010-001, California Institute of Technology,
  2010.

\bibitem{inat_benchmark}
D.~Wertheimer and B.~Hariharan.
\newblock Few-shot learning with localization in realistic settings.
\newblock In {\em IEEE Conference on Computer Vision and Pattern Recognition
  (CVPR)}, 2019.

\bibitem{wu1983convergence}
C.~J. Wu.
\newblock On the convergence properties of the em algorithm.
\newblock {\em The Annals of statistics}, pages 95--103, 1983.

\bibitem{feat}
H.-J. Ye, H.~Hu, D.-C. Zhan, and F.~Sha.
\newblock Learning embedding adaptation for few-shot learning.
\newblock In {\em IEEE Conference on Computer Vision and Pattern Recognition
  (CVPR)}, 2020.

\bibitem{zangwill1969nonlinear}
W.~I. Zangwill.
\newblock {\em Nonlinear programming: a unified approach}, volume~52.
\newblock Prentice-Hall Englewood Cliffs, NJ, 1969.

\bibitem{variational_fsl}
J.~Zhang, C.~Zhao, B.~Ni, M.~Xu, and X.~Yang.
\newblock Variational few-shot learning.
\newblock In {\em International Conference on Computer Vision (ICCV)}, 2019.

\bibitem{Laplacian}
I.~M. Ziko, J.~Dolz, E.~Granger, and I.~B. Ayed.
\newblock Laplacian regularized few-shot learning.
\newblock In {\em International Conference on Machine Learning (ICML)}, 2020.

\end{thebibliography}

\clearpage
\appendices
\section{Proof of Proposition \ref{prop:mi_theory}}
    
    \subsection{Preliminary results}
    	We first derive some results that will be needed in the main proof:
    	
    	\begin{lemma}\label{lemma:soft_vs_hard}
    		(Soft-classifier vs hard-classifier) The following relations holds:
    		\begin{align}
    			\left| \mathcal{I}(Y;\widehat{Y}) - \mathcal{I}(Y;\widehat{y}^*(X)) \right| \leq \delta(P_\Delta), 
    		\end{align}
    		where $P_\Delta = \mathbb P(\widehat{Y} \neq \widehat{y}^*(X))$, and $\delta(\cdot )$ is a strictly increasing function on the restricted domain $\left[0, \frac{|\mathcal{Y}| - 1}{|\mathcal{Y}} \right]$. 
    	\end{lemma}
    
    	\begin{IEEEproof}
    		\begin{align*}
    			\mathcal{I}(Y;\widehat{Y}) - \mathcal{I}(Y;\widehat{y}^*(X)) &= \mathcal{H}(Y|\widehat{y}^*(X)) - \mathcal{H}(Y|\widehat{Y}) \\
    			& \leq \mathcal{H}(Y|\widehat{y}^*(X)) - \mathcal{H}(Y|\widehat{Y}, \widehat{y}^*(X)) \\
    			& = \mathcal{I}(Y; \widehat{Y}|\widehat{y}^*(X)).
    		\end{align*}
    		Now, let us introduce the error variable $E= \mathbbm 1\big\{ \widehat{Y} \neq \widehat{y}^*(X) \big\}$. Then, we have:
    		\begin{align}
    			\mathcal{I}\big(Y;\widehat{Y}|\widehat{y}^*(X)\big) \leq \mathcal{I}\big((Y,E);\widehat{Y}|\widehat{y}^*(X)\big)
    		\end{align}
    		because the variable $(Y,E)$ can only contain more information than $Y$ alone. Using the chain rule for conditional mutual information, we can write:
    		\begin{align*}
    			\mathcal{I}((Y,E);\widehat{Y}|\widehat{y}^*(X)) &=
    			                 ~\mathcal{I}(E;\widehat{Y}|\widehat{y}^*(X)) 
    			                 + \mathcal{I}(Y;\widehat{Y}|\widehat{y}^*(X), E) \\
    						     &=~\mathcal{H}(E|\widehat{y}^*(X)) 
    						     - \underbrace{\mathcal{H}(E|\widehat{y}^*(X), \widehat{Y})}_{=0} \\
    						     &+ \mathbb P(E=1)~\mathcal{I}(Y;\widehat{Y}|\widehat{y}^*(X), E=1) \\
    						     &+ \mathbb P(E=0)~\underbrace{\mathcal{I}(Y;\widehat{Y}|\widehat{y}^*(X), E=0)}_{=0} \\
    						     &= \mathcal{H}(E|\widehat{y}^*(X))  \\
    						     &+ \mathbb P(E=1)~\mathcal{I}(Y;\widehat{Y}|\widehat{y}^*(X), E=1) \\
    						     & \leq \mathcal{H}(E) 
    						     + \mathbb P(E=1)~\log(|\mathcal{Y}|-1).
    		\end{align*}
    		For ease, let us note ${\mathbb P(E=1) = P_\Delta}$. Then, one can verify that $\delta: P_\Delta \rightarrow \mathcal{H}_2(P_\Delta)  + P_\Delta~\log(|\mathcal{Y}|-1)$, where ${\mathcal{H}_2(p) = -p \log (p) - (1-p) \log (1-p)}$ is the binary entropy,  is a strictly increasing function in the domain $\left[0, \frac{|\mathcal{Y}| - 1}{|\mathcal{Y}} \right]$. We can prove the second inequality:
    		\begin{align}
    			\mathcal{I}(Y;\widehat{y}^*(X)) - \mathcal{I}(Y;\widehat{Y}) \leq \delta(P_\Delta)
    		\end{align}
    		with similar arguments.
    	\end{IEEEproof}
    		
    	\begin{lemma}[Continuity of entropy \cite{csiszar2011}] \label{continuity-lemma}
            For any arbitrary discrete random variables $Y$ and $\widehat{Y}$ with probability distributions $P_{Y}$ and  $P_{\widehat{Y}}$, respectively, it follows that
            \begin{align*}
                \left| \mathcal{H}(Y) - \mathcal{H}(\widehat{Y}) \right|  \leq & ~
                \frac{1}{2} \| \P_{Y} - \P_{\widehat{Y}} \|_1 \cdot \log(| \mathcal{Y} | -1) \nonumber\\
                &+ \mathcal{H}_2\left(\frac{1}{2} \| \P_{Y} - \P_{\widehat{Y}} \|_1 \right),
            \end{align*}
            where $\| \cdot  \|_1$ denotes  the total variation distance 
            $$\| \P_{Y} - \P_{\widehat{Y}} \|_1 \coloneqq \sum\limits_{y \in \mathcal{Y}} |\P_{Y}(y) - \P_{\widehat{Y}}(y)|.$$ 
        \end{lemma}

    	\begin{lemma}\label{lemma:mi_theory_interm_prop}
    		Let us consider a soft classifier $\widehat{Y}|X$. Let us assume this classifier has a diagonal dominant confusion matrix. Then the following result holds:
    		\begin{align*}
    			\mathbb P_e \leq \frac{1}{2} \left[\mathcal{H}(Y) - \mathcal{I}(Y;\widehat{Y}) + \delta(P_\Delta) \right], 
    		\end{align*}
    		where $P_\Delta = \mathbb P(\{\widehat{Y} \neq \widehat{y}^*(X)\})$, and $\delta(\cdot )$ is a strictly increasing function on the restricted domain $\left[0, 1 - 1/ |\mathcal{Y}| \right]$. 
    	\end{lemma}
    
    	\begin{IEEEproof}
    		We will start by using a result from \cite{kovalevsky1968}, that relates the conditional entropy to the MAP error probability:
    		
    		\begin{equation*}
    			\mathcal{H}(Y|\widehat{y}^*(X)) \geq \phi^*\left(\mathbb P_e^{\text{MAP}}\right), 
    		\end{equation*}
    		
    		where $\phi^*$ is a piecewise linear convex function, and $\mathbb P_e^{\text{MAP}}$ is the error probability of the optimal MAP estimator of $Y$ given $\widehat{y}^*(X)$, i.e.:
    		
    		\begin{align}
    		    f^{\text{MAP}}(\widehat{y}^\star(x)) \coloneqq \arg \max\limits_{y\in \mathcal{Y}} \; \mathbb{P}_{Y|\widehat{y}^\star(X)}(y|\widehat{y}^\star(x)). 
    		\end{align}
    		
    		In other words, for a given sample $(x, y)$, this estimator is the best one at guessing the value of $y$ given $\widehat{y}^*(x)$ only. Note that there is a difference a \emph{prior} between $\mathbb P_e$ and $\mathbb P_e^{\text{MAP}}$. Still, they are not completely unrelated:
    		\begin{itemize}
    		 	\item First, it always holds that $\mathbb P_e^{\text{MAP}} \leq \mathbb P_e$. This can be easily seen because the identity estimator $f(\widehat{y}^*(X))=\widehat{y}^*(X)$ already achieves $\mathbb P_e$ error. Therefore, the best estimator can only be equal or better. Intuitively, the MAP estimator allows for "a posterior" correction of the $\widehat{y}^*(X)$ predictions.
    		 	\item Second, it can be shown that if the confusion matrix of the classifier $y^\star$ is diagonal dominant, i.e for any $y \neq y^\prime$:
    		 	\begin{equation} \label{eq:diagonal_dominance}
    		 		\mathbb P(\widehat{y}^*(X)=y, Y=y) \geq \mathbb P(\widehat{y}^*(X)=y, Y=y^\prime)
    		 	\end{equation}
    		 	then $\mathbb P_e^{\text{MAP}} = \mathbb P_e$. \eqref{eq:diagonal_dominance} is exactly what we initially assumed such that we can write $\mathbb P_e^{\text{MAP}} = \mathbb P_e$ in the rest of the proof.
    		\end{itemize}
    		Furthermore, if we consider the common case where $\mathbb P_e \leq 1/2$, then $\phi^*(\mathbb P_e) = 2\mathbb P_e$. Putting it all together, we finally obtain that:
    		
    		\begin{align}
    			\mathbb P_e \leq \frac{1}{2}~\mathcal{H}(Y|\widehat{y}^*(X)) = \frac{1}{2}~\left[\mathcal{H}(Y) - \mathcal{I}(Y;\widehat{y}^*(X))\right]. 
    		\end{align}
    
    		Finally, we need to relate $\mathcal{I}(Y;\widehat{y}^*(X))$ and $\mathcal{I}(Y;\widehat{Y})$. In order to do this, we use Lemma \ref{lemma:soft_vs_hard}, which allows us to write that:
    		
    		\begin{align}
    			\mathcal{I}(Y;\widehat{y}^*(X)) \geq \mathcal{I}(Y;\widehat{Y}) - \delta(P_\Delta)
    		\end{align}
    		where $P_\Delta = \mathbb P(\{ \widehat{Y} \neq \widehat{y}^*(X)\})$  reflects the uncertainty of the soft-decision (for a very peaked soft-decision, $\widehat{Y} = \widehat{y}^*(X)$ with high probability).
    		Therefore, we have shown that:
    		\begin{align}
    			\mathbb P_e \leq \frac{1}{2} \Big[ \underbrace{\mathcal{H}(Y) + \delta(P_\Delta)}_{> \mathcal{I}(Y;\widehat{Y})} - \mathcal{I}(Y;\widehat{Y}) \Big]. 
    		\end{align}
    	\end{IEEEproof}
    	
    \subsection{Proof}
        We now derive the proof of Proposition \ref{prop:mi_theory}:
    	\begin{IEEEproof}
    
            In Lemma \ref{lemma:mi_theory_interm_prop}, we showed that: 
            
            \begin{align} 
                \mathbb{P}_e \leq \frac{1}{2} \left[ \mathcal{H}(Y) -  \mathcal{I}(Y;\widehat{Y})+\delta(P_\Delta) \right].  \label{eq-upper-prop-appendix}
            \end{align}
            
            We begin by upper bounding the mutual information and the entropy as follows:
            
            \begin{align}
                \mathcal{H}(Y)-\mathcal{I}(Y;\widehat{Y}) & = \mathcal{H}(Y) - \mathcal{H}(\widehat{Y}) + \mathcal{H}(\widehat{Y}|Y)  \nonumber \\
                &\leq  \left| \mathcal{H}(Y) - \mathcal{H}(\widehat{Y}) \right| + \mathcal{H}(\widehat{Y}|Y),\label{eq-loss-1}
            \end{align}
            where \eqref{eq-loss-1} follows by the data processing inequality.  
            We now bound the absolute difference in \eqref{eq-loss-1} as follows:
            
            \begin{equation} 
                \left| \mathcal{H}(Y) - \mathcal{H}(\widehat{Y}) \right|  \leq 
                \delta \left(\sqrt{\frac{\ln 2}{2} \mathcal{D}_{\text{KL}} \big (\widehat{Y}  \| Y\big)}\right),
            \end{equation}
            
            where $P_Y$ is the prior probability distribution on the labels and $P_{\widehat{Y}} $ is the marginal probability on the labels computed from the data distribution. In order to show this, consider the following chain of inequalities: 
            
            \begin{align}
                \left| \mathcal{H}(Y) - \mathcal{H}(\widehat{Y}) \right| & \leq
                         ~\frac{1}{2} \| \P_{Y} - \P_{\widehat{Y}} \|_1 \cdot \log(| \mathcal{Y} | -1) \nonumber \\
                        &+ \mathcal{H}_2\left(\frac{1}{2} \| \P_{Y} - \P_{\widehat{Y}} \|_1 \right)\label{eq-first} \\
                        &= \delta \left(\frac{1}{2} \| \P_{Y} - \P_{\widehat{Y}} \|_1 \right), \nonumber \\
                        & \leq \delta \left(\sqrt{\frac{\ln 2}{2} \mathcal{D}_{\text{KL}} \big (\widehat{Y}  \| Y\big)}\right), \label{eq-last}
            \end{align}
            where $\| \cdot  \|_1$ denotes  the total variation distance, i.e., ${\| \P_{Y} - \P_{\widehat{Y}} \|_1 }$; and \eqref{eq-first} follows from the continuity Lemma~\ref{continuity-lemma} and \eqref{eq-last} follows from Pinsker's inequality \cite[Problem 3.18]{csiszar2011} which implies that 
            \begin{equation}
                \| \P_{Y} - \P_{\widehat{Y}} \|_1  \leq \sqrt{2 \ln(2) \cdot \mathcal{D}_{\text{KL}} \big (\widehat{Y}  \| Y\big)}.
             \end{equation}
            Provided that $\sqrt{(\ln 2) \mathcal{D}_{\text{KL}} \big (\widehat{Y}  \| Y\big) /2} \leq (| \mathcal{Y} |-1)/| \mathcal{Y} |$, by combining expressions \eqref{eq-last} and \eqref{eq-loss-1}, we have shown that 
            \begin{equation} 
                \mathcal{H}(Y)-\mathcal{I}(Y;\widehat{Y})  \leq \delta \left(\sqrt{\frac{\ln 2}{2} \mathcal{D}_{\text{KL}} \big ( \widehat{Y}  \| Y \big)}\right)
                + \mathcal{H}(\widehat{Y}|Y). \label{eq-necesito}
            \end{equation}
            We now bound the conditional entropy $\mathcal{H}(\widehat{Y}|Y)$ in \eqref{eq-necesito}. We've assumed that $\exists ~\epsilon > 0$ such that:
            \begin{align}
                \P_{\widehat{Y}|Y}(\widehat{Y}=y|Y=y) \geq 1-\epsilon, \ \ ~ \forall ~ y.    
            \end{align}
            Such relation allows us to have a tighter upper bound on  $\mathcal{H}(\widehat{Y}|Y)$ than the naive $\log(K)$:
            \begin{align}
                \mathcal{H}(\widehat{Y}|Y) =&
                    - \sum_y \P_{Y}(y) \sum_{y^\prime} \P_{\widehat{Y}|Y}(y^\prime|y) \log \P_{\widehat{Y}|Y}(y^\prime|y) \nonumber\\
                    =& - \sum_y \P_{Y}(y) \underbrace{\P_{\widehat{Y}|Y}(y|y)}_{\leq 1}  \log \underbrace{\P_{\widehat{Y}|Y}(y|y)}_{\geq 1 - \epsilon}  \nonumber\\
                     & - \sum_y \P_{Y}(y) \sum_{y^\prime \neq y} \underbrace{-\P_{\widehat{Y}|Y}(y^\prime|y) \log \P_{\widehat{Y}|Y}(y^\prime|y)}_{\leq - \epsilon \log \epsilon} \nonumber \\
                    \leq& -\log(1-\epsilon) - (|\mathcal{Y}|-1)~\epsilon~\log(\epsilon) \nonumber\\ 
                     \coloneqq  &  \, g(\epsilon).\label{eq:cond-ent}
            \end{align}
            One can check that as the confusion matrix becomes perfectly diagonal (i.e.,  $\epsilon \rightarrow 0$), $g(\epsilon)$ goes to 0.
  
            It remains to upper bound the uncertainty of the soft-classifier denoted by $P_\Delta$. To this end, we begin by observing:
            
            \begin{align} 
                P_\Delta &\coloneqq  \P(\widehat{Y} \ne \widehat{y}^\star(X)) \nonumber\\
                & = 1 - \P(\widehat{Y} = \widehat{y}^\star(X)) \nonumber\\
                & = 1 - \E_{X\widehat{Y}}[ \one \{ \widehat{Y} = \widehat{y}^\star(X) \}] \nonumber\\
                & = 1 - \E_X[\P_{\widehat{Y}|X}(\widehat{y}^\star(X)|X)]. \label{eq-nose1}
            \end{align}
            
            At this point, we recall that $\widehat{y}^*(x)  \coloneqq \argmax_y \P_{\widehat{Y}|X}(y|x)$ such that:
            
            \begin{equation} 
                \P_{\widehat{Y}|X}(\widehat{y}^\star(x)|x) \geq \P_{\widehat{Y}|X}(y|x),~ \forall (x,y) \in \mathcal{X} \times \mathcal{Y}.
            \end{equation}
            
            Since this holds for every pairs $(x,y)\in\mathcal{X}\times \mathcal{Y}$, it holds in expectation that:
            
            \begin{equation}
                \E_X[\P_{\widehat{Y}|X}(\widehat{y}^\star(X)|X)] \geq \E_X\left [ \E_{\widehat{Y}|X}[ \P_{\widehat{Y}|X}(\widehat{Y}|X) | X ]\right], \label{eq-nose2}
            \end{equation}
            
            with equality if $\widehat{Y} = \widehat{y}^\star(X)$ \emph{almost surely.} We notice that by Jensen's inequality, 
            
            \begin{align} 
                \E_X\left [ \E_{\widehat{Y}|X}[ \P_{\widehat{Y}|X}(\widehat{Y}|X) | X ]\right] & = \E_{X\widehat{Y}} \left [\exp \left( \log \P_{\widehat{Y}|X}(\widehat{Y}|X) \right)\right ] \nonumber\\ 
                & \geq \exp \left( \E_{X\widehat{Y}} \big[\log \P_{\widehat{Y}|X}(\widehat{Y}|X)\big] \right) \nonumber\\ 
                & = \exp \left( -\mathcal{H}(\widehat{Y}|X) \right),\label{eq-nose3}
            \end{align}
            
            where equality holds if $P_{\widehat{Y}|X}$ is degenerate. From expressions \eqref{eq-nose1} to  \eqref{eq-nose3}, we have:
            
            \begin{align}
                P_\Delta &\leq 1 - \exp \left( -\mathcal{H}(\widehat{Y}|X) \right) \nonumber \\
                & \leq \mathcal{H}(\widehat{Y}|X).\label{eq-nose4}
            \end{align}
            
            Therefore, and again provided that $\mathcal{H}(\widehat{Y}|X) \leq (|\mathcal{Y}| - 1) / |\mathcal{Y}|$, we can write
            
            \begin{align}
                \delta(P_\Delta) &\leq \delta(\mathcal{H}(\widehat{Y}|X)).  \label{eq:p_delta}
            \end{align}
            
            Finally, by combining \eqref{eq-upper-prop-appendix}, \eqref{eq-necesito}, \eqref{eq:cond-ent} and \eqref{eq:p_delta} together, we obtain:
            
            \begin{align*} 
                \P_e \leq & ~
                    \delta \left(\sqrt{\frac{\ln 2}{2} \mathcal{D}_{\text{KL}} \big ( \widehat{Y} \| Y\big)}\right) + \delta\left(\mathcal{H}(\widehat{Y}|X)\right) + g(\epsilon). 
            \end{align*}
            
            
            
             
            
            
            
    
        \end{IEEEproof}

\section{Proof of Proposition \ref{prop:mi_adm}}\label{app:proof_mi_adm}
    \begin{proof}
        Let us start from the initial optimization problem:
        \begin{align}\label{eq:adm_proof_1}
            \min_{\W}\quad 
            & \sum_{k=1}^K \hat{p_{k}} \log \hat{p_{k}} 
            - \frac{\alpha}{|\mathcal{Q}|} \sum_{i \in \mathcal{Q}} \sum_{k=1}^K p_{ik} \log p_{ik}  \\
            &- \frac{\lambda }{|\mathcal{S}|}\sum_{i \in \mathcal{S}}\sum_{k=1}^K y_{ik} \log p_{ik}.
        \end{align}
        We can reformulate problem (\ref{eq:adm_proof_1}) using the ADM approach, i.e., by introducing auxiliary variables $\q=[q_{ik}] \in \mathbb{R}^{|\mathcal{Q}| \times K}$ and enforcing 
        equality constraint $\q = \p$, with $\p=[p_{ik}] \in \mathbb{R}^{|\mathcal{Q}| \times K}$, in addition to pointwise simplex constraints: 
        \begin{align}\label{eq:adm_proof_2}
            \min_{\W, \q}\quad 
            & \sum_{k=1}^K \hat{q_{k}} \log \hat{q_{k}} - \frac{\alpha}{|\mathcal{Q}|} \sum_{i \in \mathcal{Q}} \sum_{k=1}^K q_{ik} \log p_{ik} \\ 
            &- \frac{\lambda }{|\mathcal{S}|}\sum_{i \in \mathcal{S}}\sum_{k=1}^K y_{ik} \log p_{ik} \nonumber \\
            \text{s.t.}\quad & q_{ik}=p_{ik}, \quad i \in \mathcal{Q}, \quad k \in \{1, \dots, K\} \nonumber \\ 
            & \sum_{k=1}^K q_{ik}=1, \quad i \in \mathcal{Q} \nonumber \\
            & q_{ik} \geq 0, \quad i \in \mathcal{Q}, \quad k \in \{1,\dots,K\}
         \end{align}
         We can solve constrained problem (\ref{eq:adm_proof_2}) with a penalty-based approach, which encourages auxiliary pointwise predictions $\q_{i}=[q_{i1}, \dots, q_{iK}]$ to be close to our model's posteriors $\p_{i}=[p_{i1}, \dots, p_{iK}]$. To add a penalty encouraging equality constraints $\q_i = \p_i$, we use the Kullback–Leibler (KL) divergence, which is given by:
         
        \begin{equation}\label{eq:penalty}
            \mathcal{D}_{\mbox{\tiny KL}}(\q_i\|\p_i) = \sum_{k=1}^K q_{ik} \log \frac{q_{ik}}{p_{ik}}.
        \end{equation}
         Thus, our constrained optimization problem becomes:
         \begin{align}\label{eq:adm_proof_3}
            \min_{\W, \q}\quad 
            & \sum_{k=1}^K \hat{q_{k}} \log \hat{q_{k}} 
            - \frac{\alpha}{|\mathcal{Q}|} \sum_{i \in \mathcal{Q}} \sum_{k=1}^K q_{ik} \log p_{ik}  \nonumber \\
            &- \frac{\lambda }{|\mathcal{S}|}\sum_{i \in \mathcal{S}}\sum_{k=1}^K y_{ik} \log p_{ik} 
            + \frac{\beta}{|\mathcal{Q}|}\sum_{i \in \mathcal{Q}}  \mathcal{D}_{\mbox{\tiny KL}}(\q_i\|\p_i),  \nonumber \\
            \text{such that}\quad & \sum_{k=1}^K q_{ik}=1, \quad i \in \mathcal{Q}, \nonumber \\
            & q_{ik} \geq 0, \quad i \in \mathcal{Q}, \quad k = [1:K].
         \end{align}
         where $\beta > 0$ is the Lagrange multiplier associated with penalty \eqref{eq:penalty}. As said in the main text, we treat $\beta$ as a fixed hyperparameter in practice.
    \end{proof}

\section{Proof of Proposition \ref{prop:adm_decreases_loss}}\label{app:proof_mi_adm_solution}
	\begin{proof}
        Recall that we consider a softmax classifier over distances to weights $\W=\{\w_1, \dots, \w_K\}$. To simplify the notations, we will omit the dependence upon $\boldsymbol{\phi}$ in what follows, and write ${\z_i=\frac{f_{\boldsymbol{\phi}}(\x_i)}{\norm{f_{\boldsymbol{\phi}}(\x_i)}}}$, such that:
        \begin{align}
            p_{ik} = \frac{e^{-\frac{\tau}{2}\norm{\z_i - \w_k}^2}}{\sum_{j=1}^K e^{-\frac{\tau}{2}\norm{\z_i - \w_j}^2}}.
        \end{align}
        Without loss of generality, we use $\tau=1$ in what follows. Plugging the expression of $p_{ik}$ into Eq. \eqref{eq:mi_adm}, and grouping terms together, we get:
        \begin{align}
            \eqref{eq:mi_adm} =& \sum_{k=1}^K \hat{q_k} \log \hat{q_k} 
            - \frac{\beta+\alpha}{|\mathcal{Q}|} \sum_{i \in \mathcal{Q}} \sum_{k=1}^K q_{ik} \log p_{ik} \nonumber \\
            &- \frac{\lambda}{|\mathcal{S}|} \sum_{i \in \mathcal{S}} \sum_{k=1}^K y_{ik} \log p_{ik} 
            + \frac{\beta}{|\mathcal{Q}|} \sum_{i \in \mathcal{Q}}\sum_{k=1}^K q_{ik} \log q_{ik} \nonumber \\
            \begin{split}
                =& \sum_{k=1}^K \hat{q_k} \log \hat{q_k} \\
                &+\frac{\beta+\alpha}{2|\mathcal{Q}|} \sum_{i \in \mathcal{Q}} \sum_{k=1}^K q_{ik} \norm{\z_i - \w_k}^2 \\
                &+ \frac{\beta+\alpha}{|\mathcal{Q}|} \sum_{i \in \mathcal{Q}} \log \left (\sum_{j=1}^K e^{-\frac{1}{2}\norm{\z_i - \w_j}^2} \right ) \\
                &+ \frac{\lambda}{2|\mathcal{S}|} \sum_{i \in \mathcal{S}}\sum_{k=1}^K y_{ik} \norm{\z_i - \w_k}^2  \\
                &+ \frac{\lambda}{|\mathcal{S}|} \sum_{i \in \mathcal{S}} \log \left (\sum_{j=1}^K e^{-\frac{1}{2}\norm{\z_i - \w_j}^2} \right ) \\
                &+ \frac{\beta}{|\mathcal{Q}|} \sum_{i \in \mathcal{Q}} \sum_{k=1}^K q_{ik} \log q_{ik}.
            \end{split}\label{eq:update_proof_1}
        \end{align}
        Now, we can solve our problem approximately by alternating two sub-steps: 
        one sub-step optimizes w.r.t classifier weights $\W$ while auxiliary variables $\q$ are fixed; another sub-step fixes $\W$ and update $\q$.
        
        \begin{itemize}
             \item $\q$-update: With weights $\mathbf W$ fixed, the objective is convex w.r.t auxiliary variables $\q_i$ (sum of linear and convex functions) and the simplex constraints are affine. Therefore, one can minimize this constrained convex problem for each $\q_i$ by solving the
                Karush-Kuhn-Tucker (KKT) conditions\footnote{Note that strong duality holds since the objective is convex and the simplex constraints are affine. This
                means that the solutions of the (KKT) conditions minimize the objective.}. The KKT conditions yield closed-form solutions for both
                primal variable $\q_i$ and the dual variable (Lagrange multiplier) corresponding to simplex constraint $\sum_{j=1}^K q_{ij}=1$.
                Interestingly, the negative entropy of auxiliary variables, i.e., $\sum_{k=1}^K q_{ik} \log q_{ik}$,  which appears in the penalty term, handles implicitly non-negativity constraints $\q_i \geq 0$. In fact, this negative entropy acts as a barrier function, restricting the domain of each $\q_i$ to non-negative values, which avoids extra dual variables and Lagrangian-dual inner iterations for constraints $\q_i \geq 0$. As we will see, the closed-form solutions of the KKT conditions satisfy these non-negativity constraints, without explicitly imposing them. In addition to non-negativity, for each point $i$, we need to handle probability simplex constraints $\sum_{k=1}^K q_{ik}=1$. Let $\gamma_i \in \mathbb{R}$ denote the Lagrangian multiplier corresponding to this constraint. The KKT conditions correspond to setting the following gradient of the Lagrangian function to zero, while enforcing the simplex constraints:   

                \begin{align}
		            \frac{\partial (\ref{eq:mi_adm})}{\partial q_{ik}} =&
		            - \frac{\beta+\alpha}{|\mathcal{Q}|} \log p_{ik} 
		            + \frac{1}{|\mathcal{Q}|} (\log \hat{q_k} + 1) \nonumber \\
		            &+ \frac{\beta}{|\mathcal{Q}|}(\log q_{ik} + 1) + \gamma_i \nonumber \\
		            =& \frac{1}{|\mathcal{Q}|} \left (\log (\frac{q_{ik}^\beta \hat{q_k}}{p_{ik}^{\beta+\alpha}}) + 1 + \beta \right ) + \gamma_i.
		        \end{align}
                This yields: 
		        \begin{align}
		        \label{first-experession-qi}
		            q_{ik} = \frac{p_{ik}^{1 + \frac{\alpha}{\beta}}}{\hat{q_k}^{\frac{1}{\beta}}}e^\frac{-(\gamma_i |\mathcal{Q}|+1+\beta)}{\beta}.
		        \end{align}
		        Applying simplex constraint $\sum_{j=1}^K q_{ij}=1$ to \eqref{first-experession-qi}, Lagrange multiplier $\gamma_i$ verifies:
		        \begin{align}
		        \label{condition_wrt_gamma}
		           e^\frac{-(\gamma_i |\mathcal{Q}|+1+\beta)}{\beta} = \frac{1}{\displaystyle \sum_{j=1}^K \frac{p_{ij}^{1+\frac{\alpha}{\beta}}}{\hat{q_j}^{\frac{1}{\beta}}}}.
		        \end{align}
		        Hence, plugging \eqref{condition_wrt_gamma} in \eqref{first-experession-qi} yields:
		        \begin{align}
		        \label{second-experession-qi}
		            q_{ik} = \frac{\displaystyle \frac{p_{ik}^{1+\frac{\alpha}{\beta}}}{\hat{q_k}^{\frac{1}{\beta}}}}{\displaystyle \sum_{j=1}^K \frac{p_{ij}^{1+\frac{\alpha}{\beta}}}{\hat{q_j}^{\frac{1}{\beta}}}}.
		        \end{align}
		        Using the definition of $\hat{q_k}$, we can decouple this equation:
		        \begin{align}\label{eq:update_proof_2}
		            \hat{q_k} = \frac{1}{|\mathcal{Q}|} \sum_{i \in \mathcal{Q}} q_{ik} \propto \sum_{i \in \mathcal{Q}} \frac{p_{ik}^{1+\frac{\alpha}{\beta}}}{\hat{q_k}^{\frac{1}{\beta}}}
		        \end{align}
		        which implies:
		        \begin{align}
		            \hat{q_k} \propto \left (\sum_{i \in \mathcal{Q}} p_{ik}^{1+\frac{\alpha}{\beta}} \right )^{\frac{\beta}{1+\beta}}.
		        \end{align}
		        Plugging this back in Eq. \eqref{second-experession-qi}, we get:
		        \begin{equation}
		        \label{third-experession-qi}
		            q_{ik} \propto \frac{p_{ik}^{1+\frac{\alpha}{\beta}}}{\left (\displaystyle \sum_{i \in \mathcal{Q}} p_{ik}^{1+\frac{\alpha}{\beta}} \right )^{\frac{1}{1+\beta}}}. 
		        \end{equation}
		        Notice that $q_{ik} \geq 0$, hence the solution fulfils the positivity constraint of the original problem. Therefore, by updating $\q^{(t+1)}$ using \eqref{third-experession-qi} we can guarantee that the solution $\q^{(t+1)}$:
		        \begin{align}
		            \mathcal{L}(\W^{(t)}, \q^{(t+1)}) \leq \mathcal{L}(\W^{(t)}, \q^{(t)}).
		        \end{align}
		        
            \item $\W$-update: Without loss of generality, we derive the update for $\w_k$, $k \in \{1, \dots, K\}$. Omitting the terms that do not involve $\w_k$, Eq. \eqref{eq:update_proof_1} reads:
		        \begin{align}
		            &\underbrace{\frac{\lambda}{2|\mathcal{S}|} \sum_{i \in \mathcal{S}} y_{ik} \norm{\z_i - \w_k}^2 
		            + \frac{\beta+\alpha}{2|\mathcal{Q}|} \sum_{i \in \mathcal{Q}} q_{ik} \norm{\z_i - \w_k}^2}_{\mathcal{C}: \text{convex}} \nonumber\\
		            &+ \frac{\lambda}{|\mathcal{S}|} \underbrace{ \sum_{i \in \mathcal{S}} \log \left ( \sum_{j=1}^K e^{-\frac{1}{2}\norm{\z_i - \w_j}^2} \right )}_{\overline{\mathcal{C}}_\mathcal{S}: \text{non-convex}} 		          \nonumber \\
		            &+ \frac{\beta+\alpha}{|\mathcal{Q}|}  \underbrace{\sum_{i \in \mathcal{Q}} \log \left ( \sum_{j=1}^K e^{-\frac{1}{2}\norm{\z_i - \w_j}^2} \right )}_{\overline{\mathcal{C}}_\mathcal{Q}: \text{non-convex}}.\label{eq:ce_reg_1}
		        \end{align}
		        One can notice that objective (\ref{eq:update_proof_1}) is not convex w.r.t $\w_k$. Actually, it can be split into convex and non-convex parts as in Eq. \eqref{eq:ce_reg_1}. Thus, we cannot simply set the gradients to 0 to get the optimal $\w_k$. \\
		        
		        \textbf{Concavity of $\overline{\mathcal{C}}_\mathcal{S}$ and $\overline{\mathcal{C}}_\mathcal{Q}$:}
		        We show in what follows that in practical cases, the non-convex parts are actually concave. To see that, let us derive the hessian of $\overline{C}_\mathcal{S}$. To simplify equations, we will denote ${C_i =  \left ( \sum_{j=1}^K e^{-\frac{1}{2}\norm{\z_i - \w_j}^2} \right )}, ~ i \in \mathcal{S}$:
		        \begin{align}
		            \frac{\partial \overline{C}_\mathcal{S}}{\partial \w_k}
		            &= \sum_{i \in \mathcal{S}} \frac{e^{-\frac{1}{2}\norm{\z_i -\w_k}^2}}{C_i} (\z_i - \w_k) \nonumber\\
		            &= \sum_{i \in \mathcal{S}} p_{ik} (\z_i - \w_k). \label{eq:hessian_1}
		        \end{align}
		        Now we compute the derivative of $p_{ik}$:
		        \begin{align}
		            \frac{\partial p_{ik}}{\partial \w_k} &= 
		            \frac{C_i e^{-\frac{1}{2}\norm{\z_i -\w_k}^2} (\z_i - \w_k) - e^{-\norm{\z_i -\w_k}^2} (\z_i - \w_k)}{C_i^2} \nonumber\\
		            &= (\z_i - \w_k) (p_{ik} - p_{ik}^2) \label{eq:hessian_2}
		        \end{align}
		        Putting \eqref{eq:hessian_1} and \eqref{eq:hessian_2} together, we have:
                \begin{align}
                    &\frac{\partial^2 \overline{C}_\mathcal{S}}{\partial^2 \w_k} \
                    = \sum_{i \in \mathcal{S}} \frac{\partial p_{ik}}{\partial \w_k} (\z_i - \w_k)^T - p_{ik} \frac{\partial \w_k}{\partial \w_k} \\
                    &= \sum_{i \in \mathcal{S}} (p_{ik}^2 - p_{ik}) (\z_i - \w_k)(\z_i - \w_k)^T - p_{ik}~I_{d \times d}. \label{eq:hessian_3}
                \end{align}
                By assumption, \eqref{eq:hessian_3} is semi-definite negative, which allows us to say that $\overline{C}_\mathcal{S}$ is a concave function of $\w_k$, for all $k$. The exact same reasoning applies to $\overline{C}_\mathcal{Q}$. \\
                

		        \textbf{Concave-Convex procedure:} Given the concavity of $\overline{\mathcal{C}}_\mathcal{S}$ and $\overline{\mathcal{C}}_\mathcal{Q}$, we find ourselves in the well-known convex-concave setting. Concave-convex techniques proceed as follows: for a function in the form of a sum of a concave term and a convex term, the concave part is replaced by its first-order approximation, while the convex part is kept as is. The result forms an auxiliary bound on the function ${\W \rightarrow \mathcal{L}_{\textrm{ADM}}(\W, \q)}$. In our case, linearizing the concave part $\overline{\mathcal{C}}_\mathcal{S}$ of the objective at the current solution $\W^{(t)}$ yields:
		        \begin{align*}
		            \overline{\mathcal{C}}_\mathcal{S}(\w_k) &\leq~
		            \overline{\mathcal{C}}_\mathcal{S}(\w_k^{(t)}) + \frac{\partial \overline{\mathcal{C}}_\mathcal{S}}{\partial \w_k} (\w_k^{(t)})^T (\w_k - \w_k^{(t)}) \nonumber \\
		            & \leq \overline{\mathcal{C}}_\mathcal{S}(\w_k^{(t)})+ \sum_{i \in \mathcal{S}} p_{ik}^{(t)} (\z_i - \w_k^{(t)})^T \w_k
		        \end{align*}
		       with equality if $\w_k = \w_k^{(t)}$. Exact same thing can be done with $\overline{\mathcal{C}}_\mathcal{S}$. Therefore, the initial objective \eqref{eq:ce_reg_1} is upper bounded by:
		       \begin{align}
		           \begin{split} \label{eq:adm_auxiliary_bound}
    		           \eqref{eq:ce_reg_1} 
    		           \leq& ~\frac{\lambda}{|\mathcal{S}|} \left( \sum_{i \in \mathcal{S}} y_{ik} \frac{\norm{\z_i - \w_k}^2}{2} \right.\\
    		           &\qquad + \left. \sum_{i \in \mathcal{S}} p_{ik}^{(t)} (\z_i - \w_k^{(t)})^T \w_k \right) \\
    		           &+ \frac{\beta+\alpha}{|\mathcal{Q}|} \left( \sum_{i \in \mathcal{Q}} q_{ik} \frac{\norm{\z_i - \w_k}^2}{2}  \right .\\
    		           & \qquad \left . + \sum_{i \in \mathcal{S}} p_{ik}^{(t)} (\z_i - \w_k^{(t)})^T \w_k \right) + \text{cste} \\
    		           &= \mathcal{E}(\w_k, \q^{(t+1)})
		           \end{split}
		       \end{align}
		       with equality if $\w_k=\w_k^{(t)}$. Now the whole benefit of $\mathcal{E}$ is that it is strictly convex in $\w_k$, and its global optimum can be obtained in closed-form by simply setting its gradient to 0:
		       \begin{align} \label{eq:adm_gradients}
		            \begin{split}
		                & \frac{\partial ~ \mathcal{E}}{\partial \w_k} 
    		            = \frac{\lambda}{|\mathcal{S}|} ~ \left[\sum_{i \in \mathcal{S}} y_{ik}(\w_k - \z_i) 
    		            + p_{ik}^{(t)}(\z_i - \w_k^{(t)})\right] \ + \\ 
    		            & \frac{\beta+\alpha}{|\mathcal{Q}|} ~ \left[\sum_{i \in \mathcal{Q}} q_{ik}(\w_k - \z_i) 
    		            + p_{ik}^{(t)}(\z_i - \w_k^{(t)})\right].
		            \end{split}
		       \end{align}
		       Setting the right-hand side of \eqref{eq:adm_gradients} to 0 exactly recovers the update \eqref{eq:w-update}, and we can guarantee that the solution $\W^{(t+1)}$ improves the initial objective:
		       \begin{align*}
		            \mathcal{L}_{\textrm{ADM}}(\W^{(t+1)}, \q^{(t+1)}) & \leq \mathcal{E}(\W^{(t+1)}, \q^{(t+1)}) \\
		            & \leq \mathcal{E}(\W^{(t)}, \q^{(t+1)})\\&  = \mathcal{L}_{\textrm{ADM}}(\W^{(t)}, \q^{(t+1)}).
		       \end{align*}
		       
        \end{itemize}
    \end{proof}

\section{Details of ADM ablation}\label{sec:details_adm_ablation}
    
    In \autoref{tab:adm_ablation_details}, we provide the $\W$ and $\q$ updates for each configuration of the TIM-ADM ablation study, whose results were presented in \autoref{tab:ablation_effect_terms}. The proof for each of these updates is very similar to the proof of Proposition \ref{prop:adm_decreases_loss} detailed in \autoref{app:proof_mi_adm_solution}. Therefore, we do not detail it here.  
    
    \begin{table}[H]
        \centering
        \resizebox{!}{0.25\textwidth}{
            \begin{tabular}{ccc}
                \toprule
                 \textbf{Loss} & $\w_k$ \textbf{update} & $q_{ik}$ \textbf{update} \\
                 \midrule
                 $\mathrm{CE}$ & $\frac{\smashoperator[r]\sum_{i \in \mathcal{S}} y_{ik} \z_i }{\smashoperator[r]\sum_{i \in \mathcal{S}} y_{ik}}$ & N/A \\
                 \hline
                 $\mathrm{CE} + \hat{\mathcal{H}}(Y_\mathcal{Q}|X_\mathcal{Q})$ & - & $\propto p_{ik}^{1+\frac{\alpha}{\beta}}$ \\
                 \hline
                 $\mathrm{CE} - \hat{\mathcal{H}}(Y_\mathcal{Q})$ & - & $\propto \frac{\smashoperator[r]p_{ik}}{\left ( \smashoperator[r]\sum_{i \in \mathcal{Q}}p_{ik} \right )^{\frac{1}{1+\beta}}}$ \\
                 \hline
                 $\mathrm{CE} - \hat{\mathcal{H}}(Y_\mathcal{Q}) + \hat{\mathcal{H}}(Y_\mathcal{Q}|X_\mathcal{Q})$ & - & - \\
                 \bottomrule
            \end{tabular}
        }
        \caption{The $\bf W$ and $\q$-updates for each case of the ablation study. "-" refers to the updates in Proposition \ref{prop:adm_decreases_loss}. "NA" refers to non-applicable.}
        \label{tab:adm_ablation_details}
    \end{table}

\section{Proof of Proposition \ref{prop:tim_adm_convergence}}
    \subsection{Background}
        In this section, we try to introduce the minimal set of required elements to understand Zangwill's theory, upon which our own convergence result is based. We first introduce the central concept of point-to-set map $\psi$ which maps a point $\thetab \in \Theta$ to a set of points $\psi(\thetab) \subset \Theta$. Intuitively, $\psi$ has to be understood as representing one iteration of the algorithm considered that, from a point $\thetab^{(t)}$ in the parameter space $\Theta$ outputs a new point in the parameter space $\thetab^{(t+1)} \in \psi(\thetab^{(t)})$ from a set of (local minima) points. We hereby recall the notion of closedness, which generalizes the concept of continuity in standard point-to-point maps to point-to-set maps:
        \begin{definition}
            (Closedness)  Let's consider two converging sequences:
            \begin{align}
                \thetab^{n}  \xrightarrow[n \to \infty]{} \thetab^*\ ,  \nonumber\\
                \widetilde{\thetab}^{n} \xrightarrow[n \to \infty]{} \widetilde{\thetab}^*\ . \nonumber
            \end{align}
            Let us also assume that:
            \begin{align}\label{eq:closed_map}
                \forall n \in \mathbb N,~\widetilde{\thetab}^{n} \in \psi(\thetab^{n}).   
            \end{align}
            Then, the point-to-set map $\psi$ is said to be closed at point $\thetab^*$ if the relation \eqref{eq:closed_map} extends to the limit $n \rightarrow \infty$, i.e:
            \begin{align}
                \tilde{\thetab}^* \in \psi(\thetab^*). \nonumber
            \end{align}
            The point-to-set map $\psi$ is said to be closed on the set $\Theta$ if it is closed at every point of $\Theta$.
        \end{definition}
        We are now ready to enunciate Zangwill's theorem:
        \begin{theorem}(\cite[p.29]{zangwill1969nonlinear}) \label{theorem:zangwill}
            Consider $\Theta$ a compact set, $\Gamma$ a subset of $\Theta$, $\psi: \Theta \rightarrow \mathcal{P}(\Theta)$ a point-to-set map, and $\mathcal{L}: \thetab \rightarrow \mathbb R$ a continuous function. 
            Assume that for any $\thetab \in \Theta \setminus \Gamma$: \vspace{0.5em}\\
            (1) $\psi(\thetab)$ is nonempty and $\psi$ closed at $\thetab$, \\
            (2) $\thetab' \in \psi(\thetab) \implies \mathcal{L}(\thetab') < \mathcal{L}(\thetab)$. \vspace{0.5em}\\
            Then, any sequence $\{\thetab^{(t)}\}_{t \in \mathbb N}$ defined by $\thetab^{(t+1)} \in \psi(\thetab^{(t)})$ has all of its limit points in $\Gamma$.
        \end{theorem}
        
        As noted in \cite{sriperumbudur2009convergence}, the general idea to prove the convergence of an iterative algorithm is to properly set $\Gamma$ and $\mathcal{L}$. The natural choice for $\Gamma$ is to set it as the set of fixed points of the algorithm $\Gamma = \big\{\thetab \in \{\Theta~|~\psi(\thetab)=\{\thetab\}\}\big\}$, and $\mathcal{L}$  as the loss the algorithm minimizes. With that in mind, assumption (2) in \autoref{theorem:zangwill} simply ensures that while the algorithm has not reached stationary points $\Gamma$, the loss is strictly decreasing.
        We first expose a result from \cite{gunawardana2005convergence} that will be used in the main proof:
        \begin{lemma} \label{lemma:closedness} (Proposition 7 of Appendix A in \cite{gunawardana2005convergence})
            For a continuous function $\phi: A \times B \rightarrow B$ defined as
            \begin{align}
                \psi(\abf) = \argmin_{\bb \in B} \phi(\abf, \bb). 
            \end{align}
            Then, $\psi$ is closed at $\abf$ if $\psi(\abf)$ is nonempty.
        \end{lemma}
        
    \subsection{Proof}
        We now start the main proof of Proposition \ref{prop:tim_adm_convergence}:
        \begin{proof} 
            The idea of the proof is to apply Theorem \ref{theorem:zangwill} with the right ingredients. 
            \textbf{Definition of $\psi$: } First, we define the point-to-set map $\psi$. First, we define the point-to-set map associated to the $\q$ and $\W$ updates:
            \begin{align}
                \psi_{\q}(\W^{(t)}, \q^{(t)}) &= (\W^{(t)},\q^{(t+1)}), \\
                \psi_{\W}(\W^{(t)}, \q^{(t+1)}) &= (\W^{(t+1)},\q^{(t+1)}), 
            \end{align}
            where $\W^{(t+1)}$ and $\q^{(t+1)}$ are given by \eqref{eq:w-update} and \eqref{eq:q-update} respectively.
            Then, the point-to-set map associated to the algorithm is simply defined as a composition of the two previous:
            \begin{align}
                \psi &= \psi_{\W} \circ \psi_{\q}.
            \end{align}
            Then, we define $\Gamma$ as the set of fixed points of $\psi$:
            \begin{align}
                \Gamma = \{(\W, \q) ~:~ (\W,\q) = \psi(\W, \q) \}.
            \end{align}
            
            We now define our parameter space. Given that the loss keeps decreasing, $\{(\W^{(t)}, \q^{(t)})\}_{t \in \mathbb N}$ has to live in the following parameter space:
            \begin{align}
                \Theta =  L_0 \times \Delta_K, 
            \end{align}
            where 
            $$L_0=\{\W ~|~  \mathcal{L}_{\textrm{ADM}}(\W, \q^{0}) \leq  \mathcal{L}_{\textrm{ADM}}(\W^0, \q^{0})\}
            $$ 
            represents a sublevel set of $ \mathcal{L}_{\textrm{ADM}}(~.~, \q^{0})$. \\
    
                
            \textbf{Assumption (1): }
            Using the continuity of $\mathcal{L}$ in both $\W$ and $\q$ and Lemma \ref{lemma:closedness}, the closedness of $\psi_{\W}$ and $\psi_{\q}$ follows. \\
                
            \textbf{Assumption (2):} We now prove that the loss strictly decreases for nonstationnary points. Consider iteration ${t \in \mathbb{N}}$, and $\W^{(t+1)},\q^{(t+1)} = \psi(\W^{(t)},\q^{(t)})$, we want to show that:
            \begin{align}   \label{eq:assumption_2}
                \begin{split}
                    (\W^{(t+1)},\q^{(t+1)}) &\neq (\W^{(t)},\q^{(t)})\\
                    &\Downarrow \\
                    \mathcal{L}(\W^{(t+1)},\q^{(t+1)}) &< \mathcal{L}(\W^{(t)},\q^{(t)}).    
                \end{split}
            \end{align}
            Let us prove the contrapose of \eqref{eq:assumption_2}, i.e let us consider $\mathcal{L}(\W^{(t+1)},\q^{(t+1)}) \leq \mathcal{L}(\W^{(t)},\q^{(t)})$. First, we can rule out the case $\mathcal{L}(\W^{(t+1)},\q^{(t+1)}) > \mathcal{L}(\W^{(t)},\q^{(t)})$ because both $\psi_{\q}$ and $\psi_{\W}$ decrease the loss. Therefore, the only possible case is that $\mathcal{L}(\W^{(t+1)},\q^{(t+1)}) = \mathcal{L}(\W^{(t)},\q^{(t)})$.
            Note that $\mathcal{L}(\W, \q)$ is strictly convex in $\q$. Therefore, 
            \begin{align}
                \mathcal{L}(\W^{(t)}, \q^{(t+1)}) < \mathcal{L}(\W^{(t)}, \q^{(t)}) \quad \text{or} \quad \q^{(t+1)} = \q^{(t)}. 
            \end{align}
            In other words, either $\q^{(t)}$ is already the global optima of $\mathcal{L}(\W^{(t)},~.~)$, and then $\q^{(t+1)} = \q^{(t)}$, or it is not and $\q^{(t+1)}$ will achieve a strictly better loss. Given that we assumed $\mathcal{L}(\W^{(t+1)},\q^{(t+1)}) = \mathcal{L}(\W^{(t)},\q^{(t)})$, and given the subsequent $\W$-update cannot increase the loss, there is no choice but ${\q^{(t+1)} = \q^{(t)}}$. Now note that $\psi_{\W}$ is a deterministic step that only depends on $\q^{(t+1)}$, we are left with ${(\W^{(t+1)}, \q^{(t+1)}) = \psi_{\W}(\W^{(t)}, \q^{(t+1)})}$, which ends proving the contrapose of \eqref{eq:assumption_2}, and achieves the full proof.
            
                    
        \end{proof}

\begin{figure}[t]
    \centering
    \begin{tabular}{c|c}
        anti-aliasing=False & anti-aliasing=True \\
        \includegraphics[width=0.47\textwidth]{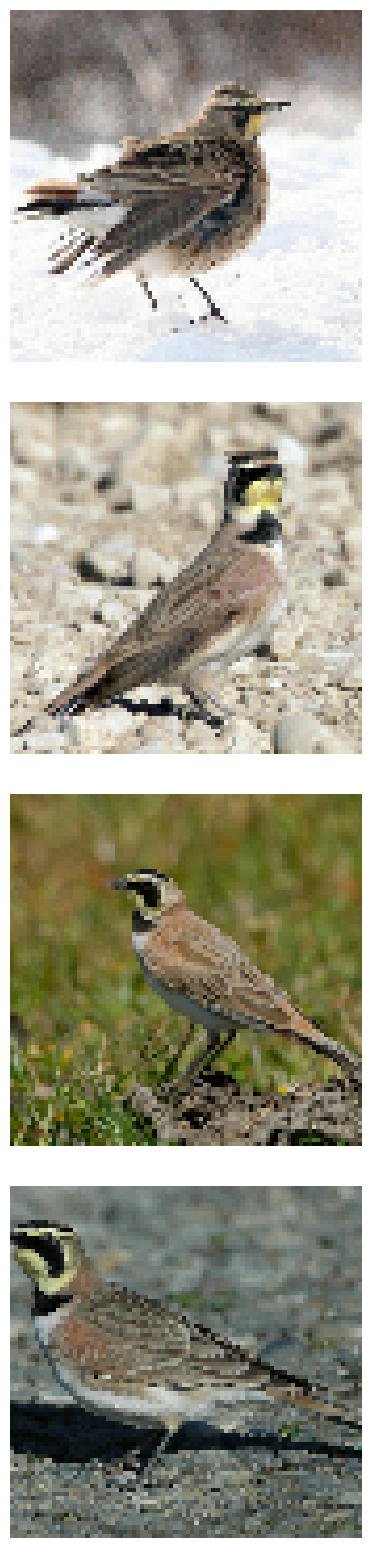} &         \includegraphics[width=0.47\textwidth]{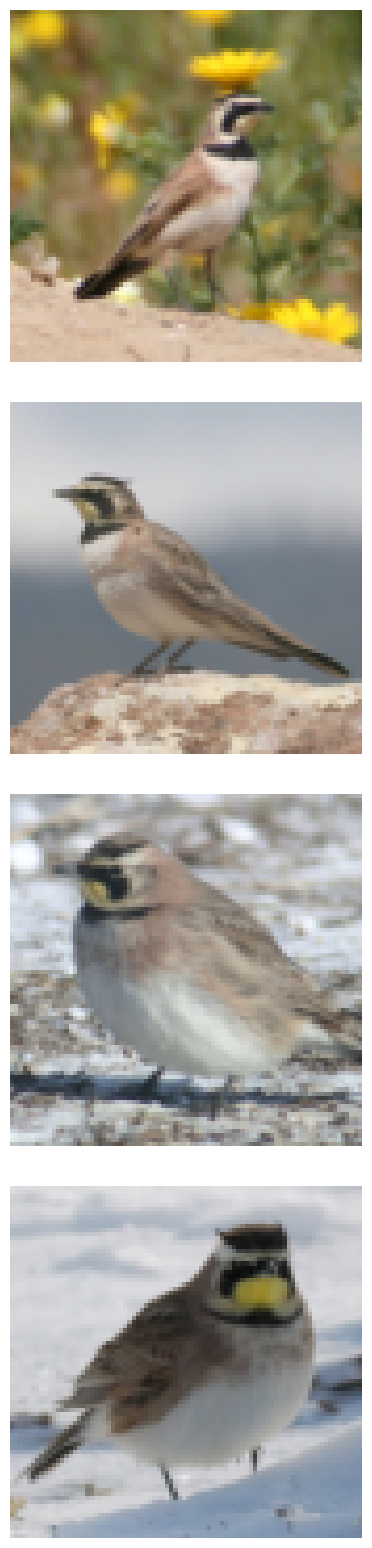}
    \end{tabular}
    \caption{Visual inspection of resized bird images from CUB \cite{cub} with and without anti-aliasing. Left pictures tend to have a more pixelated and  less smooth appearance than those on the right. While this difference may look subtle to our eye, it certainly is not for a network.}
    \label{fig:anti-aliasing}
\end{figure}
    
\section{Anti-aliasing for META-DATASET}

    In our experimental section, we showed that our PyTorch implementation of META-DATASET \cite{triantafillou2019meta} yielded significant gains over the original implementation. We found this to be caused by a simple but important implementation detail: the resize transform. In particular, two elements of the resizing have different defaults behaviors between TensorFlow and PyTorch frameworks:
    \begin{itemize}
        \item PyTorch resizes an image with dimension (H, W) to a fixed size R by multiplying both dimensions by $\max\big\{\frac{R}{H}, \frac{R}{W}\big\}$, which we can then complement by a central crop to obtain an RxR image with a preserved aspect ratio. In contrast, TensorFlow resizes the image by scaling with a factor $\min\big\{\frac{R}{H}, \frac{R}{W}\big\}$ and padding the rest with zeros.
        \item More importantly, PyTorch uses by default anti-aliasing in its resize function, while TensorFlow does not. This typically leads to seemingly more pixelated images, which can lead to significant differences on datasets where tiny details matter a lot (for instance the beak of a bird in CUB). A visual illustration of this phenomenon is presented on \autoref{fig:anti-aliasing}.
    \end{itemize}

\end{document}